\documentclass{article}

\PassOptionsToPackage{numbers, compress, sort}{natbib}
\usepackage[main, final]{neurips_2026}

\usepackage[utf8]{inputenc} %
\usepackage[T1]{fontenc}    %
\usepackage{hyperref}       %
\usepackage{url}            %
\usepackage{booktabs}       %
\usepackage{amsfonts}       %
\usepackage{nicefrac}       %
\usepackage{microtype}      %
\usepackage{xcolor}         %
\usepackage{graphicx, wrapfig, subfig} %
\usepackage{booktabs, multicol, multirow, xltabular, colortbl} %
\usepackage{amsmath, amsfonts, amssymb, nicefrac} %
\usepackage[capitalize,noabbrev]{cleveref} %

\usepackage{algorithm,algpseudocode}
\newtheorem{definition}{Definition}

\hypersetup{
  pdftitle={Efficient Test-time Adaptation through Candidate Verification and Divergence Shifts},
  pdfauthor={Seungmin Oh, Seunghun Kang, Jongbin Ryu},
  pdfsubject={Training-free test-time adaptation of vision-language models via candidate verification with divergence shifts},
  pdfkeywords={test-time adaptation, vision-language models, candidate verification, divergence shift},
}

\hypersetup{
colorlinks=true,
breaklinks=true,
citecolor={blue!50!black},
linkcolor={red!80!black},
urlcolor=magenta,
}

\title{Efficient Test-time Adaptation through Candidate Verification and Divergence Shifts}

\author{
Seungmin Oh\\
Ajou University\\
\texttt{seungminoh@ajou.ac.kr}
\And
Seunghun Kang\\
Ajou University\\
\texttt{seunghun.kang@ajou.ac.kr}
\And
Jongbin Ryu\thanks{Corresponding author.}\\
Ajou University\\
\texttt{jongbinryu@ajou.ac.kr}
}

\graphicspath{{assets/}}

\newcommand{\figMain}{
\begin{figure*}[!t]
\centering
\includegraphics[width=0.98\linewidth]{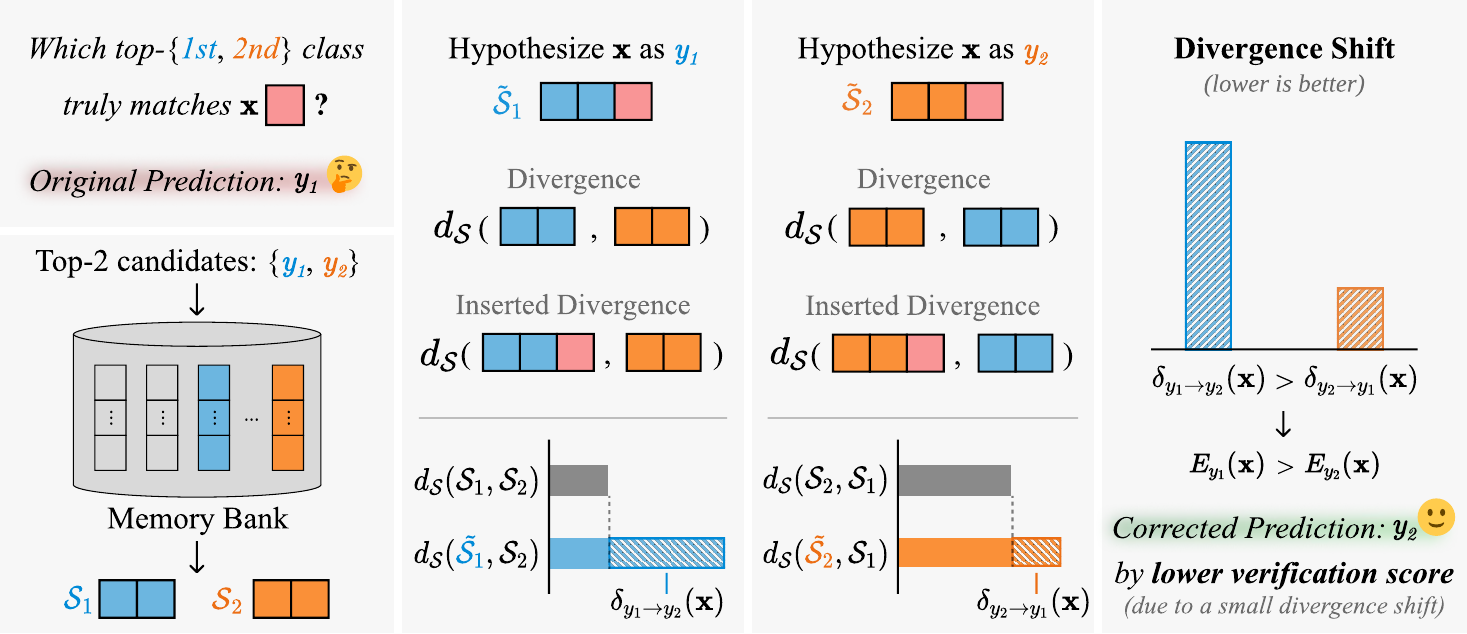}
\caption{
Overview of the Test-Time Correction (TTC) based on divergence shift estimation. TTC first selects the top-2 candidate labels, denoted by $y_1$ and $y_2$. Each candidate label is then treated as a hypothesis, and TTC estimates the divergence shifts, $\delta_{y_1 \to y_2}(\mathbf{x})$ and $\delta_{y_2 \to y_1}(\mathbf{x})$, induced by inserting the test feature $\mathbf{x}$ into the corresponding candidate subspace. Although the original top-1 prediction is $y_1$, TTC corrects the prediction to $y_2$ because inserting $\mathbf{x} $ into $\mathcal{S}_{y_2}$ induces a smaller divergence shift, yielding the lowest aggregated verification score.
}
\label{fig:overview}
\end{figure*}
}

\newcommand{\figFewshot}{
\begin{figure}[!t]
    \centering
    \subfloat[Average]{\includegraphics[width=0.24\linewidth]{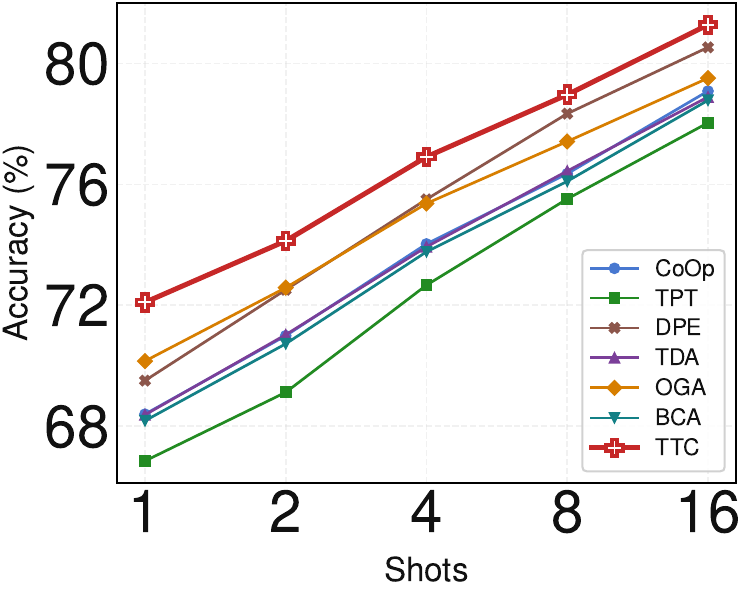}\label{fig:fewshot-average}}
    \hfill
    \subfloat[ImageNet]{\includegraphics[width=0.24\linewidth]{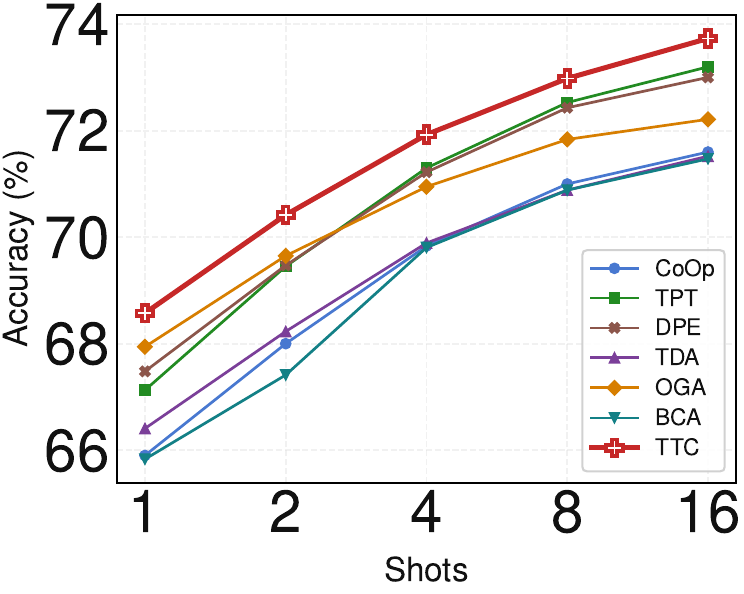}\label{fig:fewshot-imagenet}}
    \hfill
    \subfloat[Aircraft]{\includegraphics[width=0.24\linewidth]{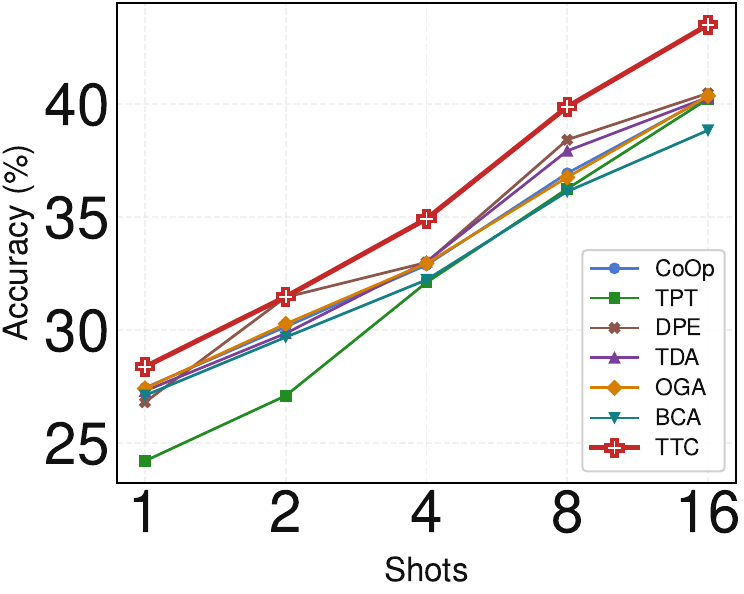}\label{fig:fewshot-aircraft}}
    \hfill
    \subfloat[Flowers102]{\includegraphics[width=0.24\linewidth]{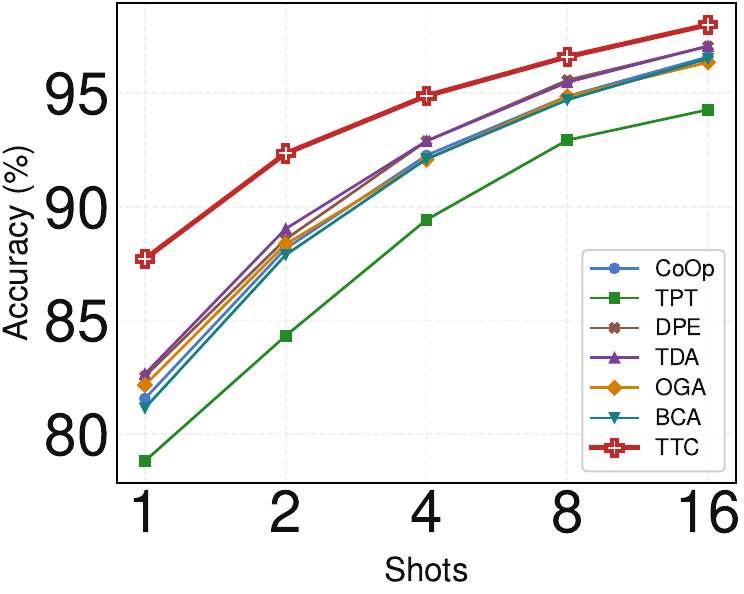}\label{fig:fewshot-flowers}}
\caption{Experimental results of few-shot classification using CoOp trained with a specified shot.}
\label{fig:fewshot}
\end{figure}
}

\newcommand{\figFewshotFull}{
\begin{figure}[H]
\centering
\includegraphics[width=0.95\linewidth]{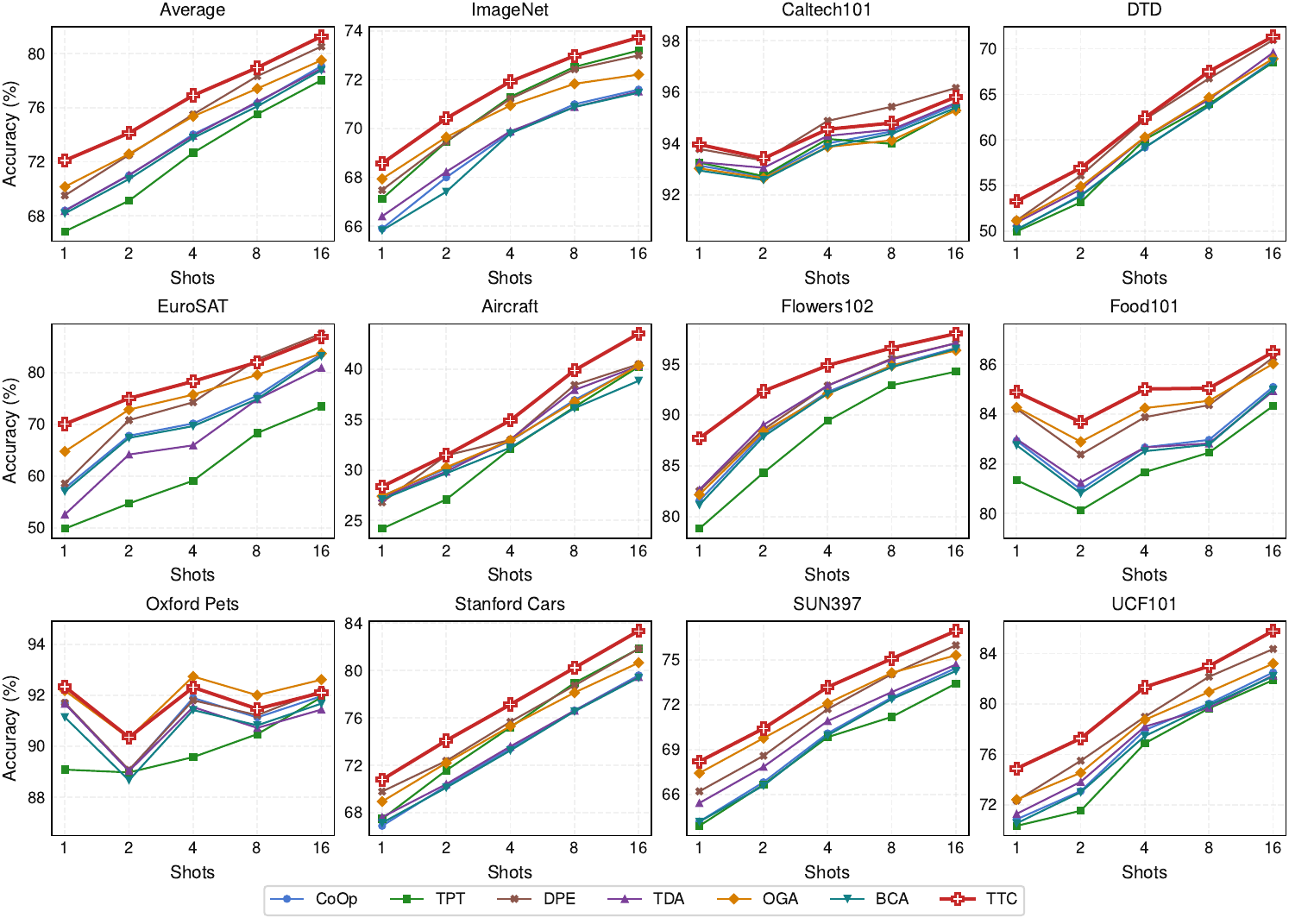}
\caption{Additional experimental results of the few-shot classification based on CoOp.}
\label{fig:fewshot-full}
\end{figure}
}

\newcommand{\figAblation}{
\begin{figure}[!t]
    \centering
    \subfloat[Top-$k$]{\includegraphics[width=0.24\linewidth]{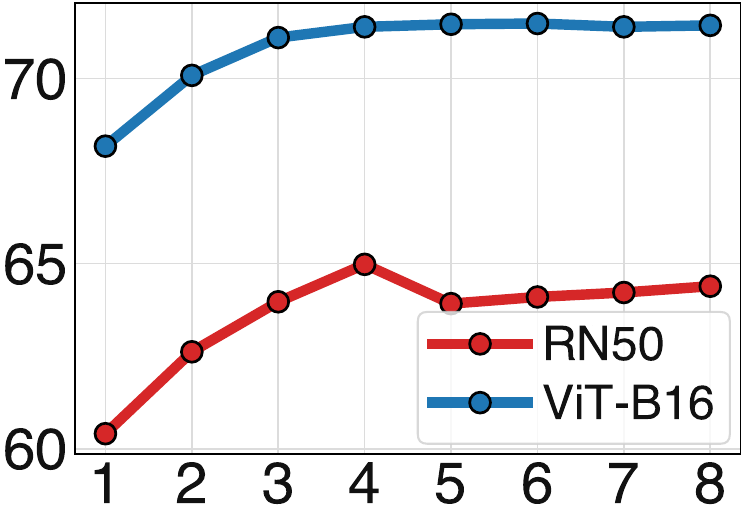}\label{fig:topk}}
    \hfill
    \subfloat[Memory Size (INet)]{\includegraphics[width=0.24\linewidth]{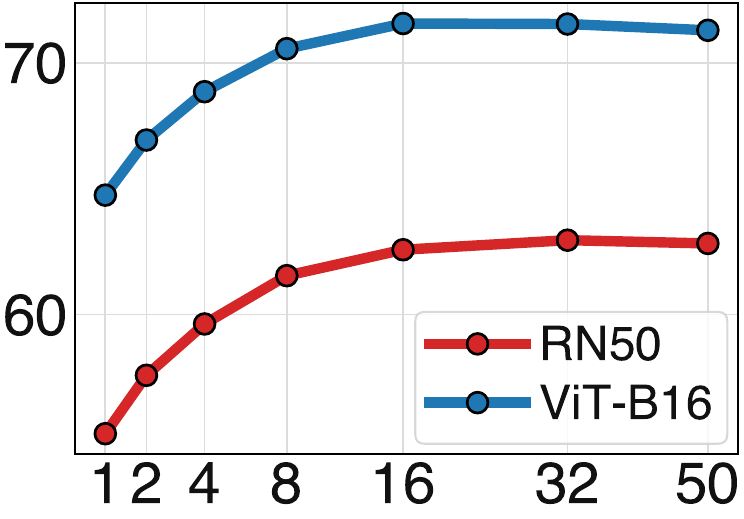}\label{fig:memsize-inet}}
    \hfill
    \subfloat[Memory Size]{\includegraphics[width=0.24\linewidth]{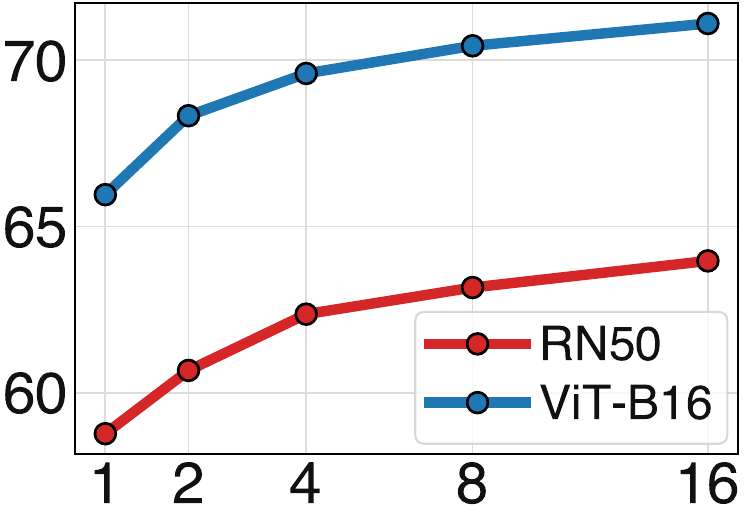}\label{fig:memsize}}
    \hfill
    \subfloat[Memory Purity]{\includegraphics[width=0.24\linewidth]{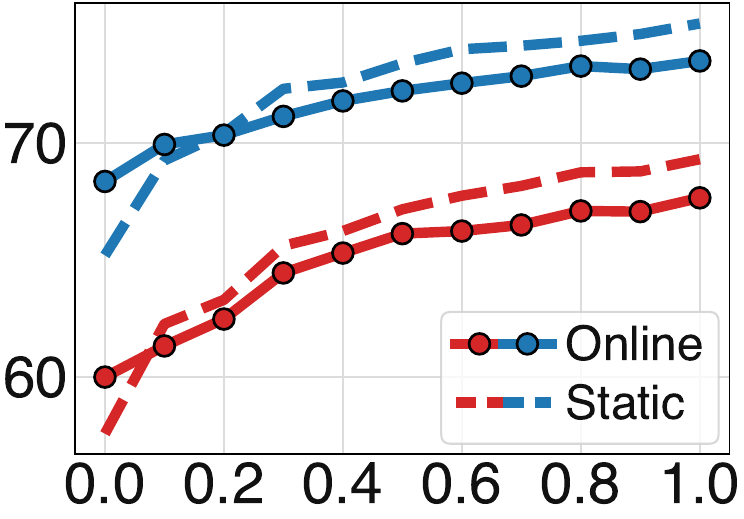}\label{fig:memacc}}
\caption{Ablation studies on top-$k$ and memory under the zero-shot classification setting.}
\label{fig:ablation}
\end{figure}
}

\newcommand{\figMemoryRecovery}{
\begin{figure}[!htbp]
\centering
    \hspace*{\fill}
    \subfloat[ResNet50]{\includegraphics[width=0.38\linewidth]{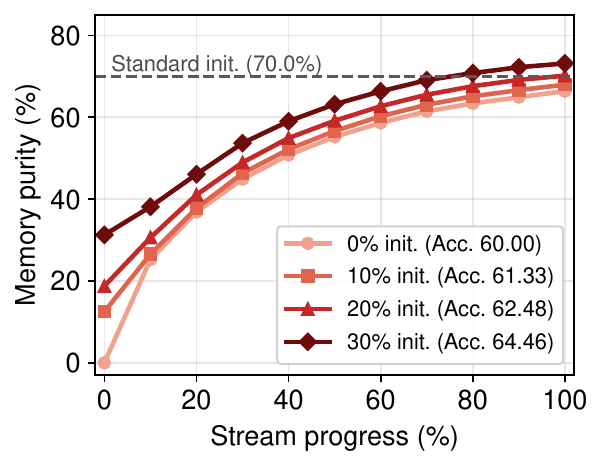}\label{fig:memory-recovery-rn50}}
    \hfill
    \subfloat[ViT-B/16]{\includegraphics[width=0.38\linewidth]{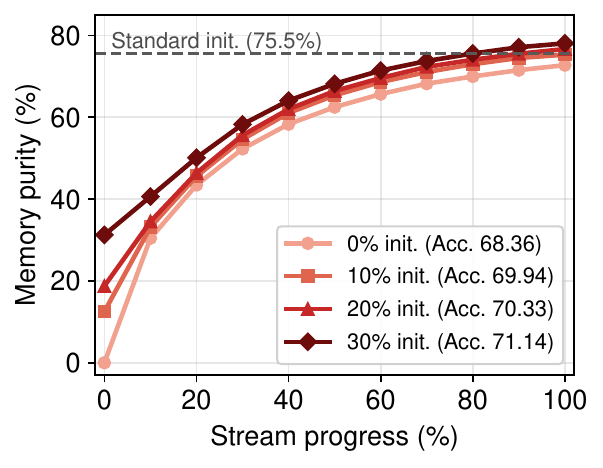}\label{fig:memory-recovery-vit}}
    \hspace*{\fill}
\caption{Memory purity along the stream under contaminated memory initialization, macro-averaged over the 11 zero-shot datasets. The legend gives the final accuracy of each setting, and the dashed line marks the initial purity of the standard initialization.}
\label{fig:memory-recovery}
\end{figure}
}

\usepackage{pifont}
\usepackage{makecell}

\newcommand{\STAB}[1]{\begin{tabular}{@{}c@{}}#1\end{tabular}}
\newcommand{\ourrow}{\rowcolor{lightgray!30}}
\definecolor{gray}{rgb}{0.502,0.502,0.502}

\newcommand{\tabZeroShot}{%
\begin{table}[t]
\centering
\caption{Experimental results of the zero-shot classification task with downstream datasets. The check mark (`\checkmark') denotes the training-free methods.}
\label{tab:zeroshot}
\resizebox{\linewidth}{!}{
\begin{tabular}{lcccccccccccc}
\toprule
\phantom{\checkmark}\ Model & Caltech & DTD & ESAT & Aircraft & Flower & Food & INet & Pets & Cars & SUN & UCF & Avg. \\
\midrule
\checkmark\ ResNet50\textsubscript{\textit{(ICML'21)}} & 83.68 & 41.68 & 27.69 & 17.40 & 63.14 & 74.95 & 59.90 & 80.79 & 58.33 & 59.53 & 55.74 & 56.62 \\
\phantom{\checkmark}\ TPT\textsubscript{\textit{(NeurIPS'22)}} & 87.02 & 40.84 & 28.33 & 17.58 & 62.69 & 74.88 & 60.74 & 84.49 & 58.46 & 61.46 & 60.82 & 57.94 \\
\phantom{\checkmark}\ DiffTPT\textsubscript{\textit{(ICCV'23)}} & 86.89 & 40.72 & 41.04 & 17.60 & 63.53 & \textbf{79.21} & 60.80 & 83.40 & 60.71 & 62.72 & 62.67 & 59.94 \\
\phantom{\checkmark}\ C-TPT\textsubscript{\textit{(ICLR'24)}} & 86.90 & 42.20 & 27.80 & 17.00 & 65.20 & 74.70 & 60.20 & 84.10 & 56.50 & 61.00 & 59.70 & 57.75 \\
\phantom{\checkmark}\ DPE\textsubscript{\textit{(NeurIPS'24)}} & 90.83 & 50.18 & 41.67 & 19.80 & 67.60 & 77.83 & 61.54 & 85.97 & 59.26 & 64.23 & 61.98 & 61.90 \\
\checkmark\ TDA\textsubscript{\textit{(CVPR'24)}} & 89.70 & 43.74 & 42.11 & 17.61 & 68.74 & 77.75 & 61.35 & 86.18 & 57.78 & 62.53 & 64.18 & 61.06 \\
\phantom{\checkmark}\ O-TPT\textsubscript{\textit{(CVPR'25)}} & 86.90 & 41.90 & 28.40 & 16.80 & 65.60 & 74.20 & 59.00 & 83.40 & 56.40 & 60.90 & 58.80 & 57.48 \\
\checkmark\ OGA\textsubscript{\textit{(CVPR'25)}} & 87.91 & 43.09 & 43.73 & 16.86 & 66.67 & 76.24 & 59.65 & \textbf{86.73} & 58.50 & 62.72 & 64.37 & 60.59 \\
\checkmark\ BCA\textsubscript{\textit{(CVPR'25)}} & 89.70 & 48.58 & 42.12 & 19.89 & 66.30 & 77.19 & 61.81 & 85.58 & 58.13 & 63.38 & 63.51 & 61.47 \\
\ourrow \checkmark\ TTC (Ours) & \textbf{91.08} & \textbf{53.72} & \textbf{47.91} & \textbf{21.09} & \textbf{71.78} & 77.86 & \textbf{62.58} & 86.48 & \textbf{60.73} & \textbf{65.53} & \textbf{65.03} & \textbf{63.98} \\
\midrule
\checkmark\ ViT-B/16\textsubscript{\textit{(ICML'21)}} & 90.89 & 44.87 & 47.01 & 24.40 & 67.65 & 85.35 & 68.50 & 86.20 & 67.60 & 64.65 & 64.36 & 64.68 \\
\phantom{\checkmark}\ TPT\textsubscript{\textit{(NeurIPS'22)}} & 94.16 & 47.75 & 42.44 & 24.78 & 68.98 & 84.67 & 68.98 & 87.79 & 66.87 & 65.50 & 68.04 & 65.45 \\
\phantom{\checkmark}\ DiffTPT\textsubscript{\textit{(ICCV'23)}} & 92.49 & 47.00 & 43.13 & 25.60 & 70.10 & \textbf{87.23} & 70.30 & 88.20 & 67.01 & 65.74 & 68.22 & 65.91 \\
\phantom{\checkmark}\ C-TPT\textsubscript{\textit{(ICLR'24)}} & 93.60 & 46.00 & 43.20 & 24.90 & 69.80 & 83.70 & 68.50 & 88.20 & 65.80 & 64.80 & 65.70 & 64.93 \\
\checkmark\ ZERO\textsubscript{\textit{(NeurIPS'24)}} & 93.66 & 46.12 & 34.33 & 25.21 & 67.68 & 86.53 & 69.31 & 87.75 & 68.04 & 65.03 & 67.77 & 64.68 \\
\phantom{\checkmark}\ DPE\textsubscript{\textit{(NeurIPS'24)}} & 94.81 & 54.20 & 55.79 & \textbf{28.95} & 75.07 & 86.17 & 69.91 & 91.14 & 67.31 & 70.07 & 70.44 & 69.44 \\
\checkmark\ TDA\textsubscript{\textit{(CVPR'24)}} & 94.24 & 47.40 & 58.00 & 23.91 & 71.42 & 86.14 & 69.51 & 88.63 & 67.28 & 67.62 & 70.66 & 67.71 \\
\phantom{\checkmark}\ TPS\textsubscript{\textit{(WACV'25)}} & \textbf{95.09} & 50.47 & 44.48 & 26.34 & 71.54 & 85.23 & 71.45 & 87.35 & 69.06 & 68.98 & 71.00 & 67.36 \\
\phantom{\checkmark}\ O-TPT\textsubscript{\textit{(CVPR'25)}} & 94.00 & 45.70 & 42.80 & 23.60 & 70.10 & 84.10 & 67.30 & 88.00 & 64.50 & 64.20 & 64.20 & 64.41 \\
\checkmark\ OGA\textsubscript{\textit{(CVPR'25)}} & 93.20 & 45.80 & \textbf{64.50} & 25.30 & 72.70 & 86.10 & 68.50 & \textbf{91.70} & 67.80 & 66.00 & 71.60 & 68.47 \\
\checkmark\ BCA\textsubscript{\textit{(CVPR'25)}} & 94.69 & 53.49 & 56.63 & 28.59 & 73.12 & 85.97 & 70.22 & 90.43 & 66.86 & 68.41 & 67.59 & 68.73 \\
\ourrow \checkmark\ TTC (Ours) & 94.24 & \textbf{55.50} & 60.81 & 28.41 & \textbf{76.61} & 86.62 & \textbf{71.57} & 91.69 & \textbf{69.99} & \textbf{71.43} & \textbf{75.26} & \textbf{71.10} \\
\bottomrule
\end{tabular}}
\end{table}
}

\newcommand{\tabDomainGen}{%
\begin{table}[t]
\centering
\caption{Experimental results of the domain generalization task. `Dist.' denotes the average accuracy of distribution-shift datasets. The check mark (`\checkmark') denotes the training-free methods.}
\label{tab:domaingen}
\small
\begin{tabular}{lcccccccc}
\toprule
\phantom{\checkmark}\ Model & \# of Aug. & INet & A & R & Ske. & V2 & Dist. & Avg. \\
\midrule
\checkmark\ ResNet50\textsubscript{\textit{(ICML'21)}} & 0 & 59.90 & 22.80 & 60.60 & 35.51 & 52.86 & 42.94 & 46.33 \\
\checkmark\ OGA\textsubscript{\textit{(CVPR'25)}} & 0 & 59.65 & 22.71 & 59.73 & 36.54 & 51.02 & 42.50 & 45.93 \\
\ourrow \checkmark\ TTC (Ours) & 0 & 62.58 & 23.44 & 61.40 & 38.85 & 54.85 & 44.64 & 48.23 \\
\phantom{\checkmark}\ TPT\textsubscript{\textit{(NeurIPS'22)}} & 63 & 60.74 & 26.67 & 59.11 & 35.09 & 54.70 & 43.89 & 47.26 \\
\phantom{\checkmark}\ DiffTPT\textsubscript{\textit{(ICCV'23)}} & 63 & 60.80 & \textbf{31.06} & 58.80 & 37.10 & 55.80 & 45.69 & 48.71 \\
\phantom{\checkmark}\ C-TPT\textsubscript{\textit{(ICLR'24)}} & 63 & 60.20 & 27.00 & 59.70 & 35.70 & 54.80 & 44.30 & 47.48 \\
\phantom{\checkmark}\ DPE\textsubscript{\textit{(NeurIPS'24)}} & 63 & 61.54 & 30.15 & \textbf{63.72} & 40.03 & 56.72 & 47.66 & 50.43 \\
\checkmark\ TDA\textsubscript{\textit{(CVPR'24)}} & 63 & 61.35 & 30.30 & 62.60 & 38.10 & 55.50 & 46.63 & 49.57 \\
\phantom{\checkmark}\ O-TPT\textsubscript{\textit{(CVPR'25)}} & 63 & 59.00 & 23.07 & 54.47 & 33.98 & 53.11 & 41.16 & 44.73 \\
\checkmark\ BCA\textsubscript{\textit{(CVPR'25)}} & 63 & 61.81 & 30.35 & 62.89 & 38.08 & 56.58 & 46.98 & 49.94 \\
\ourrow \checkmark\ TTC (Ours) & 10 & \textbf{62.58} & 30.67 & 63.30 & \textbf{41.25} & \textbf{56.91} & \textbf{48.03} & \textbf{50.94} \\
\midrule
\checkmark\ ViT-B/16\textsubscript{\textit{(ICML'21)}} & 0 & 68.50 & 49.91 & 77.64 & 48.24 & 61.89 & 59.42 & 61.24 \\
\checkmark\ OGA\textsubscript{\textit{(CVPR'25)}} & 0 & 68.50 & 50.63 & 78.15 & 50.45 & 61.62 & 60.21 & 61.87 \\
\ourrow \checkmark\ TTC (Ours) & 0 & 71.57 & 51.85 & 78.64 & 52.35 & 64.18 & 61.76 & 63.72 \\
\phantom{\checkmark}\ TPT\textsubscript{\textit{(NeurIPS'22)}} & 63 & 68.98 & 53.67 & 73.90 & 46.40 & 64.30 & 59.57 & 61.45 \\
\phantom{\checkmark}\ DiffTPT\textsubscript{\textit{(ICCV'23)}} & 63 & 70.30 & 55.68 & 75.00 & 46.80 & 65.10 & 60.65 & 62.58 \\
\phantom{\checkmark}\ C-TPT\textsubscript{\textit{(ICLR'24)}} & 63 & 68.50 & 52.90 & 78.00 & 48.50 & 63.40 & 60.70 & 62.26 \\
\checkmark\ ZERO\textsubscript{\textit{(NeurIPS'24)}} & 63 & 69.31 & 61.35 & 77.28 & 48.29 & 64.13 & 62.76 & 64.07 \\
\phantom{\checkmark}\ DPE\textsubscript{\textit{(NeurIPS'24)}} & 63 & 69.91 & 59.63 & 80.40 & 52.26 & 65.44 & 64.43 & 65.53 \\
\checkmark\ TDA\textsubscript{\textit{(CVPR'24)}} & 63 & 69.51 & 60.10 & 80.20 & 50.50 & 64.70 & 63.88 & 65.00 \\
\phantom{\checkmark}\ TPS\textsubscript{\textit{(WACV'25)}} & 63 & 71.45 & 60.61 & 80.20 & 50.88 & 64.91 & 64.15 & 65.61 \\
\phantom{\checkmark}\ O-TPT\textsubscript{\textit{(CVPR'25)}} & 63 & 67.30 & 49.87 & 72.55 & 47.12 & 61.65 & 57.80 & 59.70 \\
\checkmark\ BCA\textsubscript{\textit{(CVPR'25)}} & 63 & 70.22 & 61.14 & 80.72 & 50.87 & 64.90 & 64.41 & 65.57 \\
\ourrow \checkmark\ TTC (Ours) & 10 & \textbf{71.57} & \textbf{61.81} & \textbf{80.93} & \textbf{54.09} & \textbf{66.07} & \textbf{65.73} & \textbf{66.90} \\
\bottomrule
\end{tabular}
\end{table}
}

\newcommand{\tabBaseToNovelAndCrossDataset}{%
\begin{table}[t]
\centering
\caption{Experimental results of the base-to-novel generalization task using a CoOp-tuned ViT-B/16. The results are the harmonic mean of base- and novel-class accuracies.}
\label{tab:base2novel}
\resizebox{\linewidth}{!}{%
\begin{tabular}{lcccccccccccc}
\toprule
Model & Caltech & DTD & ESAT & Aircraft & Flower & Food & INet & Pets & Cars & SUN & UCF & Avg. \\
\midrule
CoOp & 93.36 & 57.83 & 69.55 & 31.57 & 75.45 & 85.57 & 71.88 & 94.33 & 67.93 & 71.13 & 69.48 & 72.32 \\
\;\;+ TPT & 93.71 & 57.92 & 60.62 & 31.82 & 74.81 & 84.47 & 73.67 & 93.95 & 69.07 & 71.51 & 70.43 & 71.74 \\
\;\;+ DPE & \textbf{94.56} & 59.80 & 64.06 & 31.54 & 76.29 & 86.81 & 73.51 & 94.37 & 70.30 & 72.58 & 71.52 & 73.08 \\
\;\;+ TDA & 94.32 & 58.91 & 66.50 & 31.89 & 76.59 & 85.80 & 72.09 & 94.02 & 69.39 & 72.29 & 70.73 & 72.70 \\
\;\;+ OGA & 93.14 & 59.16 & \textbf{76.75} & 31.91 & 76.60 & \textbf{87.99} & 73.33 & \textbf{94.88} & 69.01 & 74.11 & 72.08 & 74.07 \\
\;\;+ BCA & 93.33 & 57.74 & 67.75 & 30.93 & 75.32 & 85.46 & 71.76 & 93.99 & 68.01 & 71.11 & 69.36 & 72.04 \\
\ourrow \;\;+ TTC & 94.23 & \textbf{61.38} & 73.89 & \textbf{32.72} & \textbf{78.02} & 87.95 & \textbf{74.30} & 94.82 & \textbf{72.71} & \textbf{75.15} & \textbf{72.87} & \textbf{75.01} \\
\bottomrule
\end{tabular}}
\vspace{\floatsep}
\caption{Experimental results of the cross-dataset evaluation task. We perform zero-shot prediction using a CoOp-tuned ViT-B/16 trained with 16 shots per class on ImageNet.}
\label{tab:crossdataset}
\resizebox{\linewidth}{!}{%
\begin{tabular}{lc|ccccccccccc}
\toprule
Model & INet & Caltech & DTD & ESAT & Aircraft & Flower & Food & Pets & Cars & SUN & UCF & Avg. \\
\midrule
CoOp & 71.60 & 92.17 & 41.43 & 48.30 & 20.03 & 67.13 & 85.80 & 89.80 & 64.87 & 64.27 & 67.87 & 64.84 \\
\;\;+ TPT & 73.20 & 93.51 & 42.38 & 44.62 & 20.30 & 66.69 & 84.74 & 89.82 & 67.06 & 65.50 & 68.46 & 65.12 \\
\;\;+ DPE & 73.00 & 91.49 & 42.22 & 46.72 & 19.78 & 66.95 & 86.63 & 89.82 & 68.24 & 65.67 & 69.56 & 65.46 \\
\;\;+ TDA & 72.97 & 91.85 & 42.71 & 44.37 & \textbf{21.01} & 67.09 & 86.58 & 89.61 & 67.17 & 65.40 & 69.27 & 65.27 \\
\;\;+ OGA & 72.21 & 92.33 & 42.32 & \textbf{61.18} & 20.21 & 67.40 & 86.35 & 89.94 & 66.99 & 66.91 & 69.75 & 66.87 \\
\;\;+ BCA & 71.45 & 91.95 & 41.37 & 48.71 & 19.95 & 66.80 & 85.52 & 89.53 & 65.15 & 64.14 & 67.30 & 64.72 \\
\ourrow \;\;+ TTC & \textbf{73.73} & \textbf{93.87} & \textbf{43.36} & 56.38 & 20.13 & \textbf{70.39} & \textbf{86.80} & \textbf{90.13} & \textbf{68.50} & \textbf{67.64} & \textbf{72.26} & \textbf{67.56} \\
\bottomrule
\end{tabular}}
\end{table}
}

\newcommand{\tabResources}{%
\begin{table}[t]
\centering
\caption{Comparison of computing resources such as GPU and CPU memory, throughput (Thr.), wall-clock time (Wall), and accuracy (Acc.) on ImageNet zero-shot classification. All performance measurements are conducted using a single NVIDIA RTX A5000.}
\label{tab:resources}
\resizebox{\linewidth}{!}{%
\begin{tabular}{lcccccccccc}
\toprule
& \multicolumn{5}{c}{ResNet50} & \multicolumn{5}{c}{ViT-B/16} \\
\cmidrule(lr){2-6} \cmidrule(lr){7-11}
Model
  & \makecell{GPU\\(GB)$\downarrow$}
  & \makecell{CPU\\(MB)$\downarrow$}
  & \makecell{Thr.\\(img/s)$\uparrow$}
  & \makecell{Wall\\(s)$\downarrow$}
  & \makecell{Acc.\\(\%)$\uparrow$}
  & \makecell{GPU\\(GB)$\downarrow$}
  & \makecell{CPU\\(MB)$\downarrow$}
  & \makecell{Thr.\\(img/s)$\uparrow$}
  & \makecell{Wall\\(s)$\downarrow$}
  & \makecell{Acc.\\(\%)$\uparrow$}  \\
\midrule
DPE\textsubscript{\textit{(NeurIPS'24)}} & 0.38 & 1039 & 16.2 & 3080.3 & 61.54 & 0.20 & 1009 & 17.0 & 2935.5 & 69.91 \\
TDA\textsubscript{\textit{(CVPR'24)}} & 0.40 & 520 & 163.1 & 306.6 & 61.35 & 0.24 & 520 & 184.1 & 271.6 & 69.51 \\
OGA\textsubscript{\textit{(CVPR'25)}} & 2.26 & 544 & 10.6 & 4712.4 & 59.65 & 2.01 & 545 & 20.8 & 2398.7 & 68.50 \\
BCA\textsubscript{\textit{(CVPR'25)}} & 0.40 & 472 & 956.2 & 52.4 & 61.81 & 0.26 & 471 & 932.5 & 53.7 & 70.22 \\
\midrule
\ourrow TTC (Ours) & \textbf{0.31} & \textbf{142} & \textbf{1103.4} & \textbf{45.4} & \textbf{62.58} & \textbf{0.17} & \textbf{135} & \textbf{2049.0} & \textbf{24.5} & \textbf{71.57} \\
\bottomrule
\end{tabular}}
\end{table}
}

\newcommand{\tabMetricAblationZS}{
\captionof{table}{Ablation study of different divergence metrics.}
\label{tab:metric-ablation-zs}
\small\centering\setlength{\tabcolsep}{2.5pt}
\begin{tabular}{ccccccc}
\toprule
  & Metric & INet & Caltech & DTD & ESAT & Avg. \\
\midrule
\multirow{4}{*}{\STAB{\rotatebox[origin=c]{90}{ResNet50}}} & -- & 59.90 & 83.68 & 41.68 & 27.69 & 56.62 \\
 & $l_2$ dist. & 52.68 & 72.70 & 38.24 & 27.60 & 50.01 \\
 & Cos. Sim. & 61.72 & 90.63 & 51.65 & 35.63 & 61.58 \\
 & NSE & 62.58 & 91.08 & 53.72 & 47.91 & 63.98 \\
\midrule
\multirow{4}{*}{\STAB{\rotatebox[origin=c]{90}{ViT-B/16}}} & -- & 68.50 & 90.89 & 44.87 & 47.01 & 64.68 \\
 & $l_2$ dist. & 62.12 & 85.11 & 42.08 & 32.75 & 57.60 \\
 & Cos. Sim. & 70.37 & 94.56 & 55.08 & 56.69 & 69.61 \\
 & NSE & 71.57 & 94.24 & 55.50 & 60.81 & 71.10 \\
\bottomrule
\end{tabular}
}

\newcommand{\tabFunctionAblationZS}{
\captionof{table}{Ablation study of the three components of the aggregated verification score.}
\label{tab:function-ablation-zs}
\small\centering\setlength{\tabcolsep}{6pt}\renewcommand{\arraystretch}{1.05}
\begin{tabular}{ccccc}
\toprule
$d_{\mathbf{x}}$ & $\delta_{y \to r}$ & $\delta_{r \to y}$ & ResNet50 & ViT-B/16 \\
\midrule
\checkmark &  &  & 59.56 & 67.39 \\
 & \checkmark &  & 61.84 & 69.73 \\
 &  & \checkmark & 62.63 & 70.29 \\
\midrule
\checkmark & \checkmark &  & 63.60 & 70.85 \\
\checkmark &  & \checkmark & 63.73 & 71.04 \\
 & \checkmark & \checkmark & 63.25 & 70.72 \\
\midrule
\checkmark & \checkmark & \checkmark & 63.98 & 71.10 \\
\bottomrule
\end{tabular}
}

\newcommand{\tabHitRate}{%
\begin{table}[t]
\centering
\caption{Top-$3$ hit rate (Hit@3) and accuracy inside ($A_{\mathrm{in}}$) and outside ($A_{\mathrm{out}}$) the top-$3$ candidates. Hit@3 is shared by all methods in a column, and $A_{\mathrm{out}}=0$ for the initial top-$1$ and Ours (TTC).
ZS: zero-shot, DG: domain generalization, FS: few-shot, B2N: base-to-novel, Cross: cross-dataset.
}
\label{tab:topk-summary}
\resizebox{\linewidth}{!}{
\begin{tabular}{lcccccccc}
\toprule
 & \multicolumn{2}{c}{ZS} & \multicolumn{2}{c}{DG} & FS & \multicolumn{2}{c}{B2N} & Cross \\
\cmidrule(lr){2-3}\cmidrule(lr){4-5}\cmidrule(lr){7-8}
 & ResNet50 & ViT-B/16 & ResNet50 & ViT-B/16 & ViT-B/16 & Base & Novel & ViT-B/16 \\
\midrule
Hit@3 & 80.90 & 85.39 & 67.37 & 81.04 & 89.24 & 94.46 & 81.70 & 80.30 \\
\midrule
\multicolumn{9}{l}{\textit{$A_{\mathrm{in}}=P(\mathrm{correct}\mid\mathrm{GT}\in\text{top-}3)$}} \\
Initial top-$1$ & 72.62 & 78.27 & 68.45 & 76.24 & 81.41 & 86.75 & 76.94 & 77.49 \\
DPE\textsubscript{\textit{(NeurIPS'24)}} & 73.83 & 79.33 & 69.02 & 77.03 & 82.91 & 87.66 & 77.30 & 77.21 \\
TDA\textsubscript{\textit{(CVPR'24)}} & 72.78 & 77.14 & 69.06 & 76.85 & 81.26 & 86.55 & 77.68 & 77.51 \\
OGA\textsubscript{\textit{(CVPR'25)}} & 71.60 & 77.76 & 65.34 & 75.26 & 82.37 & 86.85 & 79.23 & 79.28 \\
BCA\textsubscript{\textit{(CVPR'25)}} & 73.94 & 78.93 & 68.85 & 76.55 & 81.07 & 86.41 & 76.59 & 77.24 \\
\ourrow TTC (Ours) & \textbf{77.26} & \textbf{81.77} & \textbf{69.75} & \textbf{78.14} & \textbf{84.73} & \textbf{88.51} & \textbf{81.00} & \textbf{80.96} \\
\midrule
\multicolumn{9}{l}{\textit{$A_{\mathrm{out}}=P(\mathrm{correct}\mid\mathrm{GT}\notin\text{top-}3)$}} \\
DPE\textsubscript{\textit{(NeurIPS'24)}} & 1.98 & 2.79 & 0.96 & 0.74 & 1.13 & 0.49 & 1.54 & 3.79 \\
TDA\textsubscript{\textit{(CVPR'24)}} & 1.14 & 0.84 & 0.58 & 0.64 & 0.90 & 0.40 & 0.86 & 1.24 \\
OGA\textsubscript{\textit{(CVPR'25)}} & 4.08 & 4.01 & 2.35 & 2.22 & 4.50 & 5.50 & 6.35 & 4.17 \\
BCA\textsubscript{\textit{(CVPR'25)}} & 0.08 & 0.09 & 0.01 & 0.01 & 0.03 & 0.01 & 0.00 & 0.08 \\
\bottomrule
\end{tabular}}
\end{table}
}

\newcommand{\tabTaxonomy}{%
\begin{table}[t]
\centering
\caption{Success and failure cases of TTC within the hit set, macro-averaged over the five settings with ViT-B/16. The margin is the median gap between the verification scores of the chosen candidate and the runner-up, and the pairs are representative ImageNet confusions.}
\label{tab:taxonomy}
\small
\resizebox{\linewidth}{!}{
\begin{tabular}{lcc>{\raggedright\arraybackslash}p{0.38\linewidth}}
\toprule
Case & \% of hit set & Margin & Representative pairs\\
\midrule
Success (correct) & 82.07 & 0.434 & Alaskan tundra wolf $\to$ grey wolf; standard $\to$ miniature Schnauzer \\
Failure (right $\to$ wrong) & 3.33 & 0.067 & projectile $\to$ missile; \quad\quad\quad\quad\quad\quad\quad
sunglass $\to$ sunglasses \\
Failure (wrong $\to$ wrong) & 14.60 & 0.127 & ear of corn $\to$ corn; gazelle $\to$ impala \\
\bottomrule
\end{tabular}}
\end{table}
}

\newcommand{\tabTaxonomyPerTask}{%
\begin{table}[H]
\centering
\caption{Success and failure cases of TTC within the hit set for each setting with ViT-B/16.}
\label{tab:taxonomy-per-task}
\small
\begin{tabular}{l*{6}{>{\centering\arraybackslash}p{1.45cm}}}
\toprule
 & \multicolumn{2}{c}{Success (correct)} & \multicolumn{2}{c}{Failure (right $\to$ wrong)} & \multicolumn{2}{c}{Failure (wrong $\to$ wrong)} \\
\cmidrule(lr){2-3}\cmidrule(lr){4-5}\cmidrule(lr){6-7}
Setting & \% & Margin & \% & Margin & \% & Margin \\
\midrule
ZS & 81.77 & 0.414 & 3.78 & 0.075 & 14.45 & 0.122 \\
DG & 78.14 & 0.399 & 3.57 & 0.062 & 18.29 & 0.139 \\
FS & 84.73 & 0.457 & 2.45 & 0.051 & 12.82 & 0.118 \\
B2N & 84.75 & 0.492 & 2.39 & 0.064 & 12.86 & 0.141 \\
Cross & 80.96 & 0.405 & 4.46 & 0.080 & 14.58 & 0.116 \\
\midrule
Avg. & 82.07 & 0.434 & 3.33 & 0.067 & 14.60 & 0.127 \\
\bottomrule
\end{tabular}
\end{table}
}

\newcommand{\tabHitRatePerDataset}{%
\begin{table*}[!htbp]
\centering
\caption{Per-dataset top-$3$ hit rate (Hit@3) and $A_{\mathrm{in}}=P(\mathrm{correct}\mid\mathrm{GT}\in\text{top-}3)$ of the initial top-$1$ candidate (Init.) and of Ours (TTC). FS is averaged over 1 to 16 shots and B2N over the base and novel splits, with CoOp-tuned ViT-B/16 prompts, where Init. is the CoOp prediction.}
\label{tab:topk-per-dataset}
\resizebox{\linewidth}{!}{
\begin{tabular}{lccccccccccccccc}
\toprule
 & \multicolumn{3}{c}{ZS/DG (ResNet50)} & \multicolumn{3}{c}{ZS/DG (ViT-B/16)} & \multicolumn{3}{c}{FS} & \multicolumn{3}{c}{B2N} & \multicolumn{3}{c}{Cross} \\
\cmidrule(lr){2-4}\cmidrule(lr){5-7}\cmidrule(lr){8-10}\cmidrule(lr){11-13}\cmidrule(lr){14-16}
Dataset & Hit@3 & Init. & TTC & Hit@3 & Init. & TTC & Hit@3 & Init. & TTC & Hit@3 & Init. & TTC & Hit@3 & Init. & TTC \\
\midrule
Caltech & 98.17 & 90.70 & \textbf{92.77} & 99.23 & 94.44 & \textbf{94.97} & 99.29 & 94.63 & \textbf{95.18} & 98.92 & 94.55 & \textbf{95.40} & 99.11 & 93.00 & \textbf{94.72} \\
DTD & 73.82 & 64.21 & \textbf{72.78} & 74.53 & 70.74 & \textbf{74.46} & 78.12 & 75.52 & \textbf{79.58} & 79.92 & 76.90 & \textbf{80.92} & 58.29 & 71.07 & \textbf{74.38} \\
ESAT & 75.52 & 49.45 & \textbf{63.45} & 78.78 & 68.55 & \textbf{77.20} & 91.93 & 76.92 & \textbf{85.25} & 92.56 & 78.80 & \textbf{82.78} & 75.15 & 64.27 & \textbf{75.02} \\
Aircraft & 38.97 & 50.04 & \textbf{54.12} & 51.31 & 53.68 & \textbf{55.38} & 59.37 & 56.30 & \textbf{59.70} & 57.16 & 58.08 & \textbf{60.44} & 37.76 & 53.04 & \textbf{53.31} \\
Flower & 81.97 & 80.39 & \textbf{87.57} & 85.02 & 85.72 & \textbf{90.11} & 97.57 & 92.87 & \textbf{96.21} & 88.50 & 88.82 & \textbf{91.16} & 80.77 & 83.11 & \textbf{87.15} \\
Food & 91.10 & 84.58 & \textbf{85.47} & 95.91 & 89.50 & \textbf{90.31} & 94.27 & 87.96 & \textbf{90.19} & 95.34 & 89.83 & \textbf{92.29} & 95.87 & 89.50 & \textbf{90.54} \\
INet & 81.49 & 75.25 & \textbf{76.80} & 88.08 & 79.15 & \textbf{81.25} & 87.69 & 78.98 & \textbf{81.56} & 88.99 & 80.99 & \textbf{83.65} & -- & -- & -- \\
Pets & 97.82 & 86.90 & \textbf{88.41} & 98.99 & 91.24 & \textbf{92.62} & 99.14 & 91.94 & \textbf{92.51} & 99.61 & 94.70 & \textbf{95.20} & 96.29 & 93.26 & \textbf{93.60} \\
Cars & 83.27 & 68.91 & \textbf{72.92} & 90.96 & 73.19 & \textbf{76.95} & 93.67 & 78.26 & \textbf{82.27} & 91.66 & 75.11 & \textbf{79.93} & 89.29 & 72.65 & \textbf{76.71} \\
SUN & 85.23 & 73.88 & \textbf{76.88} & 88.55 & 76.89 & \textbf{80.66} & 89.22 & 77.98 & \textbf{81.53} & 89.17 & 80.66 & \textbf{84.88} & 85.30 & 75.35 & \textbf{79.30} \\
UCF & 82.58 & 74.55 & \textbf{78.75} & 87.92 & 77.90 & \textbf{85.60} & 91.33 & 84.12 & \textbf{88.04} & 87.02 & 81.88 & \textbf{85.61} & 85.19 & 79.67 & \textbf{84.83} \\
\midrule
Avg. & 80.90 & 72.62 & \textbf{77.26} & 85.39 & 78.27 & \textbf{81.77} & 89.24 & 81.41 & \textbf{84.73} & 88.08 & 81.85 & \textbf{84.75} & 80.30 & 77.49 & \textbf{80.96} \\
\midrule
INet-A & 45.85 & \textbf{51.99} & 51.12 & 74.13 & 68.62 & \textbf{69.95} & -- & -- & -- & -- & -- & -- & -- & -- & -- \\
INet-R & 77.89 & 78.56 & \textbf{78.84} & 90.35 & 86.37 & \textbf{87.04} & -- & -- & -- & -- & -- & -- & -- & -- & -- \\
INet-Sketch & 55.84 & 65.00 & \textbf{69.58} & 69.79 & 70.77 & \textbf{75.01} & -- & -- & -- & -- & -- & -- & -- & -- & -- \\
INet-V2 & 75.76 & 71.45 & \textbf{72.40} & 82.85 & 76.28 & \textbf{77.47} & -- & -- & -- & -- & -- & -- & -- & -- & -- \\
\midrule
DG Avg. & 67.37 & 68.45 & \textbf{69.75} & 81.04 & 76.24 & \textbf{78.14} & -- & -- & -- & -- & -- & -- & -- & -- & -- \\
\bottomrule
\end{tabular}}
\end{table*}
}

\newcommand{\tabZeroShotBackbones}{%
\begin{table}[H]
\centering
\caption{Experimental results of the zero-shot classification task on CLIP ViT-B/32 and ViT-L/14.}
\label{tab:zeroshot-backbones}
\resizebox{\linewidth}{!}{%
\begin{tabular}{lcccccccccccc}
\toprule
Model & Caltech & DTD & ESAT & Aircraft & Flower & Food & INet & Pets & Cars & SUN & UCF & Avg. \\
\midrule
ViT-B/32\textsubscript{\textit{(ICML'21)}} & 92.54 & 49.47 & 51.23 & 20.61 & 66.99 & 80.23 & 64.65 & 86.56 & 61.65 & 65.82 & 64.31 & 64.01 \\
DPE\textsubscript{\textit{(NeurIPS'24)}} & 93.39 & 52.36 & 53.05 & 21.09 & 69.59 & 80.73 & 65.44 & 87.27 & 62.60 & 67.59 & 66.38 & 65.41 \\
TDA\textsubscript{\textit{(CVPR'24)}} & 92.82 & 51.30 & 54.32 & 20.64 & 69.35 & 80.45 & 65.27 & 86.84 & 61.71 & 66.93 & 65.13 & 64.98 \\
OGA\textsubscript{\textit{(CVPR'25)}} & 91.56 & 50.95 & 53.75 & 21.06 & 68.90 & 79.98 & 64.60 & 87.65 & 62.77 & 66.78 & 65.58 & 64.87 \\
BCA\textsubscript{\textit{(CVPR'25)}} & 92.33 & 50.12 & 50.86 & 21.09 & 67.11 & 80.05 & 64.76 & 86.89 & 61.72 & 66.00 & 64.66 & 64.15 \\
\ourrow TTC (Ours) & \textbf{93.79} & \textbf{52.96} & \textbf{56.74} & \textbf{21.63} & \textbf{71.99} & \textbf{81.17} & \textbf{66.00} & \textbf{88.39} & \textbf{63.60} & \textbf{68.64} & \textbf{67.51} & \textbf{66.58} \\
\midrule
ViT-L/14\textsubscript{\textit{(ICML'21)}} & 96.55 & 61.88 & 64.41 & 35.76 & 79.37 & 91.07 & 76.50 & 93.51 & 78.55 & 71.96 & 74.23 & 74.89 \\
DPE\textsubscript{\textit{(NeurIPS'24)}} & \textbf{97.04} & 64.36 & 67.20 & \textbf{38.13} & 81.73 & 91.34 & 77.15 & 93.57 & 80.28 & 73.62 & 76.34 & 76.43 \\
TDA\textsubscript{\textit{(CVPR'24)}} & 96.31 & 63.36 & 67.25 & 37.47 & 81.97 & 91.04 & 76.88 & 93.24 & 78.80 & 72.74 & 75.50 & 75.87 \\
OGA\textsubscript{\textit{(CVPR'25)}} & 96.63 & 63.06 & \textbf{68.96} & 36.99 & 82.14 & 90.81 & 76.84 & 94.09 & 80.87 & 73.19 & 76.87 & 76.40 \\
BCA\textsubscript{\textit{(CVPR'25)}} & 95.94 & 62.65 & 62.27 & 36.60 & 79.09 & 90.97 & 76.44 & 92.91 & 79.04 & 71.98 & 74.60 & 74.77 \\
\ourrow TTC (Ours) & 96.84 & \textbf{65.25} & 66.33 & 37.50 & \textbf{83.96} & \textbf{91.47} & \textbf{78.37} & \textbf{94.82} & \textbf{81.35} & \textbf{74.79} & \textbf{78.14} & \textbf{77.16} \\
\bottomrule
\end{tabular}}
\end{table}
}

\newcommand{\tabDomainGenBackbones}{%
\begin{table}[H]
\centering
\caption{Experimental results of the domain generalization task on CLIP ViT-B/32 and ViT-L/14. `Dist.' denotes the average accuracy of distribution-shift datasets.}
\label{tab:domaingen-backbones}
\small
\begin{tabular}{lcccccccc}
\toprule
Model & \# of Aug. & INet & A & R & Ske. & V2 & Dist. & Avg. \\
\midrule
ViT-B/32\textsubscript{\textit{(ICML'21)}} & 0 & 64.65 & 31.68 & 69.57 & 43.04 & 57.06 & 50.34 & 53.20 \\
OGA\textsubscript{\textit{(CVPR'25)}} & 0 & 64.60 & 32.12 & 69.73 & 44.19 & 56.53 & 50.64 & 53.43 \\
\ourrow TTC (Ours)& 0 & 66.00 & 32.15 & 70.20 & 45.70 & 57.59 & 51.41 & 54.33 \\
DPE\textsubscript{\textit{(NeurIPS'24)}} & 63 & 65.44 & \textbf{45.89} & \textbf{74.82} & 48.52 & 60.46 & \textbf{57.42} & \textbf{59.03} \\
TDA\textsubscript{\textit{(CVPR'24)}} & 63 & 65.27 & 45.71 & 74.21 & 47.58 & \textbf{60.47} & 56.99 & 58.65 \\
BCA\textsubscript{\textit{(CVPR'25)}} & 63 & 64.76 & 45.33 & 74.53 & 47.05 & 60.12 & 56.76 & 58.36 \\
\ourrow TTC (Ours) & 10 & \textbf{66.00} & 45.77 & 73.37 & \textbf{48.71} & 60.26 & 57.03 & 58.82 \\
\midrule
ViT-L/14\textsubscript{\textit{(ICML'21)}} & 0 & 76.50 & 71.91 & 88.15 & 60.37 & 70.92 & 72.84 & 73.57 \\
OGA\textsubscript{\textit{(CVPR'25)}} & 0 & 76.84 & 72.43 & 88.59 & 61.45 & 70.42 & 73.22 & 73.95 \\
\ourrow TTC (Ours) & 0 & 78.37 & 73.65 & 88.62 & 62.81 & 71.30 & 74.10 & 74.95 \\
DPE\textsubscript{\textit{(NeurIPS'24)}} & 63 & 77.15 & 80.52 & \textbf{90.73} & 63.98 & 72.26 & 76.87 & 76.93 \\
TDA\textsubscript{\textit{(CVPR'24)}} & 63 & 76.88 & 80.80 & 90.65 & 63.37 & 72.14 & 76.74 & 76.77 \\
BCA\textsubscript{\textit{(CVPR'25)}} & 63 & 76.44 & 80.59 & 90.55 & 62.83 & 72.17 & 76.53 & 76.51 \\
\ourrow TTC (Ours) & 10 & \textbf{78.37} & \textbf{81.68} & 90.34 & \textbf{64.42} & \textbf{72.38} & \textbf{77.21} & \textbf{77.44} \\
\bottomrule
\end{tabular}
\end{table}
}

\newcommand{\tabBaseToNovelFull}{%
\begin{table}[H]
\centering
\caption{Experimental results of the base-to-novel generalization task using a CoOp-tuned ViT-B/16. `H.M.' denotes the harmonic mean of base- and novel-class accuracies.}
\label{tab:base2novel-full}
\resizebox{\linewidth}{!}{%
\begin{tabular}{llcccccccccccc}
\toprule
Model & Sets & Caltech & DTD & ESAT & Aircraft & Flower & Food & INet & Pets & Cars & SUN & UCF & Avg. \\
\midrule
 & Base & 97.80 & 79.83 & 91.70 & 41.30 & 97.83 & 88.53 & 76.40 & 93.57 & 78.13 & 80.73 & 84.30 & 82.74 \\
 & Novel & 89.33 & 45.33 & 56.03 & 25.63 & 61.47 & 82.80 & 67.87 & 95.10 & 60.13 & 63.60 & 59.23 & 64.23 \\
\multirow{-3}{*}{CoOp} & H.M. & 93.36 & 57.83 & 69.55 & 31.57 & 75.45 & 85.57 & 71.88 & 94.33 & 67.93 & 71.13 & 69.48 & 72.32 \\
\midrule
 & Base & 98.28 & 80.44 & 80.26 & 40.02 & 96.20 & 85.78 & 77.88 & 93.51 & 80.38 & 79.16 & 85.33 & 81.57 \\
 & Novel & 89.56 & 45.25 & 48.70 & 26.41 & 61.21 & 83.20 & 69.88 & 94.39 & 60.56 & 65.21 & 59.96 & 64.03 \\
\multirow{-3}{*}{\;\;+ TPT} & H.M. & 93.71 & 57.92 & 60.62 & 31.82 & 74.81 & 84.47 & 73.67 & 93.95 & 69.07 & 71.51 & 70.43 & 71.74 \\
\midrule
 & Base & 98.56 & 82.37 & 92.91 & 39.80 & 98.23 & 89.51 & 77.57 & 93.85 & 80.72 & 81.46 & 86.13 & 83.74 \\
 & Novel & 90.87 & 46.94 & 48.88 & 26.11 & 62.36 & 84.28 & 69.86 & 94.89 & 62.26 & 65.45 & 61.15 & 64.82 \\
\multirow{-3}{*}{\;\;+ DPE} & H.M. & 94.56 & 59.80 & 64.06 & 31.54 & 76.29 & 86.81 & 73.51 & 94.37 & 70.30 & 72.58 & 71.52 & 73.08 \\
\midrule
 & Base & 98.00 & 80.71 & 89.79 & 40.80 & 98.23 & 88.47 & 76.29 & 93.35 & 78.20 & 80.86 & 83.90 & 82.60 \\
 & Novel & 90.90 & 46.38 & 52.80 & 26.17 & 62.77 & 83.29 & 68.32 & 94.69 & 62.36 & 65.36 & 61.13 & 64.93 \\
\multirow{-3}{*}{\;\;+ TDA} & H.M. & 94.32 & 58.91 & 66.50 & 31.89 & 76.59 & 85.80 & 72.09 & 94.02 & 69.39 & 72.29 & 70.73 & 72.70 \\
\midrule
 & Base & 97.37 & 79.40 & 91.36 & 40.90 & 97.78 & 89.65 & 76.99 & 93.78 & 78.42 & 81.51 & 85.52 & 82.97 \\
 & Novel & 89.26 & 47.14 & 66.17 & 26.15 & 62.96 & 86.39 & 69.99 & 96.01 & 61.62 & 67.93 & 62.29 & 66.90 \\
\multirow{-3}{*}{\;\;+ OGA} & H.M. & 93.14 & 59.16 & 76.75 & 31.91 & 76.60 & 87.99 & 73.33 & 94.88 & 69.01 & 74.11 & 72.08 & 74.07 \\
\midrule
 & Base & 97.87 & 79.90 & 91.44 & 39.56 & 97.75 & 88.42 & 76.24 & 93.39 & 78.13 & 80.61 & 83.94 & 82.48 \\
 & Novel & 89.23 & 45.21 & 53.90 & 25.49 & 61.32 & 82.69 & 67.78 & 94.59 & 60.26 & 63.65 & 59.22 & 63.94 \\
\multirow{-3}{*}{\;\;+ BCA} & H.M. & 93.33 & 57.74 & 67.75 & 30.93 & 75.32 & 85.46 & 71.76 & 93.99 & 68.01 & 71.11 & 69.36 & 72.04 \\
\midrule
\ourrow  & Base & 98.26 & 82.18 & 93.68 & 43.72 & 98.20 & 89.57 & 78.08 & 93.64 & 80.41 & 82.76 & 87.21 & \textbf{84.34} \\
\ourrow  & Novel & 90.54 & 48.99 & 61.15 & 26.17 & 64.80 & 86.40 & 70.87 & 96.03 & 66.44 & 68.83 & 62.66 & \textbf{67.54} \\
\ourrow \multirow{-3}{*}{\;\;+ TTC} & H.M. & 94.23 & 61.38 & 73.89 & 32.72 & 78.02 & 87.95 & 74.30 & 94.82 & 72.71 & 75.15 & 72.87 & \textbf{75.01} \\
\bottomrule
\end{tabular}}
\end{table}
}

\newcommand{\tabDatasetInfo}{
\begin{table*}[!htbp]
\caption{Detailed dataset information used in experiments. The number of training samples matches that used when training CoOp for few-shot, base-to-novel, and cross-dataset evaluation.}
\label{tab:dataset-info}
\centering
\begin{tabular}{lrrr}
\toprule
Dataset & Train Samples & Test Samples & Classes \\
\midrule
ImageNet & 16000 & 50000 & 1000 \\
ImageNet-A & -- & 7500 & 200 \\
ImageNet-R & -- & 30000 & 200 \\
ImageNet-Sketch & -- & 50889 & 1000 \\
ImageNet-V2 & -- & 10000 & 1000 \\
Caltech101 & 1600 & 2465 & 100 \\
DTD & 752 & 1692 & 47 \\
EuroSAT & 160 & 8100 & 10 \\
Aircraft & 1600 & 3333 & 100 \\
Flowers102 & 1632 & 2463 & 102 \\
Food101 & 1616 & 30300 & 101 \\
Oxford Pets & 592 & 3669 & 37 \\
Stanford Cars & 3136 & 8041 & 196 \\
SUN397 & 6352 & 19850 & 397 \\
UCF101 & 1616 & 3783 & 101 \\
\bottomrule
\end{tabular}
\end{table*}
}

\newcommand{\tabMetricAblationZSFull}{
\begin{table}[H]
\small\centering\setlength{\tabcolsep}{3pt}
\caption{Per-dataset results of the ablation study of different divergence metrics.}
\label{tab:metric-ablation-zs-full}
\centering
\begin{tabular}{cccccccccccccc}
\toprule
 & Metric & Caltech & DTD & ESAT & Aircraft & Flower & Food & INet & Pets & Cars & SUN & UCF & Avg. \\
\midrule
\multirow{4}{*}{\STAB{\rotatebox[origin=c]{90}{ResNet50}}} & -- & 83.68 & 41.68 & 27.69 & 17.40 & 63.14 & 74.95 & 59.90 & 80.79 & 58.33 & 59.53 & 55.74 & 56.62 \\
 & $l_2$ dist. & 72.70 & 38.24 & 27.60 & 14.58 & 51.93 & 72.23 & 52.68 & 81.00 & 48.90 & 50.16 & 40.10 & 50.01 \\
 & Cos. Sim. & 90.63 & 51.65 & 35.63 & 20.40 & 68.66 & 77.17 & 61.72 & 85.58 & 57.94 & 64.15 & 63.86 & 61.58 \\
 & NSE & 91.08 & 53.72 & 47.91 & 21.09 & 71.78 & 77.86 & 62.58 & 86.48 & 60.73 & 65.53 & 65.03 & 63.98 \\
\midrule
\multirow{4}{*}{\STAB{\rotatebox[origin=c]{90}{ViT-B/16}}} & -- & 90.89 & 44.87 & 47.01 & 24.40 & 67.65 & 85.35 & 68.50 & 86.20 & 67.60 & 64.65 & 64.36 & 64.68 \\
 & $l_2$ dist. & 85.11 & 42.08 & 32.75 & 22.32 & 55.26 & 83.79 & 62.12 & 87.11 & 58.09 & 55.11 & 49.83 & 57.60 \\
 & Cos. Sim. & 94.56 & 55.08 & 56.69 & 28.29 & 75.96 & 86.06 & 70.37 & 91.11 & 67.14 & 69.50 & 70.98 & 69.61 \\
 & NSE & 94.24 & 55.50 & 60.81 & 28.41 & 76.61 & 86.62 & 71.57 & 91.69 & 69.99 & 71.43 & 75.26 & 71.10 \\
\bottomrule
\end{tabular}
\end{table}
}

\newcommand{\tabFunctionAblationZSFull}{
\begin{table}[H]
\small\centering\setlength{\tabcolsep}{1.7pt}
\caption{Per-dataset results of the ablation study of the three components of the aggregated verification score.}
\label{tab:function-ablation-zs-full}
\begin{tabular}{lccccccccccccccc}
\toprule
& $d_{\mathbf{x}}$ & $\delta_{y \to r}$ & $\delta_{r \to y}$ & Caltech & DTD & ESAT & Aircraft & Flower & Food & INet & Pets & Cars & SUN & UCF & Avg. \\
\midrule
\multirow{7}{*}{\STAB{\rotatebox[origin=c]{90}{ResNet50}}} & \checkmark &  &  & 86.21 & 52.30 & 41.73 & 20.49 & 66.91 & 72.66 & 59.41 & 73.43 & 56.83 & 62.60 & 62.57 & 59.56 \\
 &  & \checkmark &  & 90.10 & 49.53 & 39.57 & 20.10 & 68.45 & 77.65 & 62.35 & 85.88 & 58.86 & 64.48 & 63.23 & 61.84 \\
 &  &  & \checkmark & 91.03 & 51.30 & 41.33 & 20.55 & 69.31 & 77.79 & 62.55 & 86.26 & 59.59 & 64.95 & 64.26 & 62.63 \\
 \cmidrule(l){2-16}
 & \checkmark & \checkmark &  & 90.99 & 52.42 & 47.59 & 21.09 & 70.81 & 77.90 & 62.81 & 86.05 & 60.12 & 65.47 & 64.31 & 63.60 \\
 & \checkmark &  & \checkmark & 90.87 & 53.01 & 47.46 & 20.82 & 71.13 & 77.89 & 62.83 & 86.07 & 60.34 & 65.54 & 65.11 & 63.73 \\
 &  & \checkmark & \checkmark & 91.12 & 52.48 & 42.79 & 21.27 & 70.20 & 77.98 & 62.80 & 86.64 & 59.97 & 65.51 & 65.00 & 63.25 \\
 \cmidrule(l){2-16}
 & \checkmark & \checkmark & \checkmark & 91.08 & 53.72 & 47.91 & 21.09 & 71.78 & 77.86 & 62.58 & 86.48 & 60.73 & 65.53 & 65.03 & 63.98 \\
\midrule
\multirow{7}{*}{\STAB{\rotatebox[origin=c]{90}{ViT-B/16}}} & \checkmark &  &  & 89.37 & 53.25 & 56.56 & 28.35 & 72.11 & 83.14 & 68.51 & 81.00 & 67.08 & 69.00 & 72.93 & 67.39 \\
 &  & \checkmark &  & 94.32 & 54.79 & 55.86 & 28.95 & 75.11 & 86.32 & 70.80 & 91.28 & 68.35 & 69.66 & 71.58 & 69.73 \\
 &  &  & \checkmark & 94.40 & 55.50 & 56.42 & 28.98 & 76.57 & 86.58 & 71.16 & 91.58 & 68.87 & 70.20 & 72.93 & 70.29 \\
 \cmidrule(l){2-16}
 & \checkmark & \checkmark &  & 94.48 & 54.73 & 61.77 & 28.44 & 75.72 & 86.76 & 71.64 & 91.33 & 69.47 & 71.13 & 73.91 & 70.85 \\
 & \checkmark &  & \checkmark & 94.52 & 55.20 & 60.67 & 28.68 & 76.21 & 86.72 & 71.66 & 91.63 & 69.84 & 71.36 & 74.99 & 71.04 \\
 &  & \checkmark & \checkmark & 94.44 & 54.79 & 58.33 & 29.34 & 76.98 & 86.65 & 71.46 & 91.77 & 69.64 & 70.66 & 73.86 & 70.72 \\
 \cmidrule(l){2-16}
 & \checkmark & \checkmark & \checkmark & 94.24 & 55.50 & 60.81 & 28.41 & 76.61 & 86.62 & 71.57 & 91.69 & 69.99 & 71.43 & 75.26 & 71.10 \\
\bottomrule
\end{tabular}
\end{table}
}

\newcommand{\tabConfidenceInterval}{
\begin{table}[H]
\small\centering\setlength{\tabcolsep}{3.4pt}
\caption{95\% confidence intervals of our experimental results over three random seeds. ZS/DG, FS, B2N, and Cross denote zero-shot classification and domain generalization, few-shot classification, base-to-novel generalization, and cross-dataset evaluation, respectively.}
\label{tab:confidence-interval}
\begin{tabular}{lccccccccccc}
\toprule
 & \multicolumn{2}{c}{ZS/DG} & \multicolumn{5}{c}{FS} & \multicolumn{3}{c}{B2N} & Cross \\
\cmidrule(lr){2-3}\cmidrule(lr){4-8}\cmidrule(lr){9-11}\cmidrule(lr){12-12}
Dataset & ResNet50 & ViT-B/16 & 1 & 2 & 4 & 8 & 16 & Base & Novel & H.M. & ViT-B/16 \\
\midrule
Caltech & 0.231 & 0.080 & 0.256 & 0.672 & 0.842 & 0.417 & 0.191 & 0.527 & 2.428 & 1.547 & 0.138 \\
DTD & 0.677 & 0.201 & 0.470 & 2.140 & 1.266 & 0.278 & 0.536 & 2.108 & 2.714 & 2.710 & 1.378 \\
ESAT & 0.500 & 3.164 & 2.724 & 3.546 & 1.102 & 3.924 & 1.167 & 0.880 & 7.051 & 5.303 & 1.910 \\
Aircraft & 0.347 & 0.278 & 1.842 & 0.641 & 2.076 & 0.706 & 0.711 & 0.745 & 2.652 & 2.298 & 1.588 \\
Flower & 0.306 & 0.096 & 1.255 & 0.351 & 1.044 & 0.253 & 0.046 & 0.598 & 5.018 & 3.721 & 2.421 \\
Food & 0.138 & 0.098 & 0.504 & 0.845 & 0.111 & 0.308 & 0.150 & 0.206 & 2.267 & 1.267 & 0.250 \\
INet & 0.140 & 0.138 & 0.301 & 0.267 & 0.115 & 0.140 & 0.261 & 0.147 & 0.602 & 0.282 & -- \\
Pets & 0.252 & 0.467 & 0.719 & 0.294 & 1.253 & 1.011 & 0.526 & 0.526 & 0.642 & 0.495 & 0.950 \\
Cars & 0.418 & 0.189 & 0.612 & 0.693 & 0.974 & 0.508 & 0.559 & 0.374 & 4.935 & 2.995 & 0.437 \\
SUN & 0.149 & 0.124 & 0.328 & 0.505 & 0.535 & 0.118 & 0.194 & 0.125 & 1.970 & 1.207 & 0.629 \\
UCF & 1.115 & 1.009 & 0.862 & 0.765 & 0.467 & 0.515 & 0.096 & 0.372 & 4.855 & 3.163 & 3.028 \\
\midrule
INet-A & 0.069 & 0.088 & -- & -- & -- & -- & -- & -- & -- & -- & -- \\
INet-R & 0.053 & 0.081 & -- & -- & -- & -- & -- & -- & -- & -- & -- \\
INet-Sketch & 0.079 & 0.091 & -- & -- & -- & -- & -- & -- & -- & -- & -- \\
INet-V2 & 0.028 & 0.205 & -- & -- & -- & -- & -- & -- & -- & -- & -- \\
\bottomrule
\end{tabular}
\end{table}
}

\newcommand{\tabDynamicK}{%
\begin{table}[!htbp]
\centering
\caption{Fixed and entropy-guided dynamic top-$k$ with ViT-B/16, averaged over the 11 zero-shot classification datasets.}
\label{tab:dynamic-k}
\small
\begin{tabular}{lccc}
\toprule
Variant & Mean $k$ & Acc. & Thr. (img/s) \\
\midrule
\ourrow Fixed $k=3$ (default) & 3.00 & 71.10 & 2049.0 (100\%) \\
Entropy, $k_{\max}=5$ & 3.47 & 71.28 & 1636 (80\%) \\
Entropy, $k_{\max}=8$ & 4.30 & 71.34 & 1174 (57\%) \\
\bottomrule
\end{tabular}
\end{table}
}

\newcommand{\tabPairedSignificance}{%
\begin{table}[H]
\centering
\caption{Paired accuracy differences between TTC and each baseline on the same three stream permutations. We report per-seed differences, their mean with the 95\% confidence interval (Gap), and each baseline's own 95\% accuracy confidence interval across seeds (Baseline CI).}
\label{tab:paired-significance}
\small
\begin{tabular}{llccccc}
\toprule
 & & \multicolumn{3}{c}{Per-seed difference} & & \\
\cmidrule(lr){3-5}
Setting & Baseline & Seed 1 & Seed 2 & Seed 3 & Gap $\pm$ 95\% CI & Baseline CI \\
\midrule
\multirow{3}{*}{ZS Avg.\ (ResNet50)} & TDA & $+2.19$ & $+1.87$ & $+2.29$ & $+2.12 \pm 0.55$ & $\pm 0.17$ \\
 & OGA & $+2.44$ & $+1.92$ & $+1.99$ & $+2.12 \pm 0.70$ & $\pm 0.74$ \\
 & BCA & $+3.83$ & $+3.60$ & $+3.91$ & $+3.78 \pm 0.41$ & $\pm 0.09$ \\
\midrule
\multirow{3}{*}{ZS Avg.\ (ViT-B/16)} & TDA & $+1.78$ & $+1.37$ & $+1.83$ & $+1.66 \pm 0.63$ & $\pm 0.40$ \\
 & OGA & $+1.60$ & $+1.57$ & $+1.70$ & $+1.62 \pm 0.17$ & $\pm 0.21$ \\
 & BCA & $+3.11$ & $+3.28$ & $+3.38$ & $+3.26 \pm 0.33$ & $\pm 0.33$ \\
\bottomrule
\end{tabular}
\end{table}
}

\newcommand{\tabMetricCost}{%
\begin{table}[!htbp]
\centering
\caption{Accuracy and throughput with regard to the divergence metric on ImageNet zero-shot task.}
\label{tab:metric-cost}
\small
\begin{tabular}{lcccc}
\toprule
 & \multicolumn{2}{c}{ResNet50} & \multicolumn{2}{c}{ViT-B/16} \\
\cmidrule(lr){2-3} \cmidrule(lr){4-5}
Metric & Acc. (\%)$\uparrow$ & Thr. (img/s)$\uparrow$ & Acc. (\%)$\uparrow$ & Thr. (img/s)$\uparrow$ \\
\midrule
Cos. Sim. & 61.72 & 1890.1 & 70.37 & 3578.8 \\
\ourrow NSE (default) & 62.58 & 1103.4 & 71.57 & 2049.0 \\
\bottomrule
\end{tabular}
\end{table}
}

\newcommand{\tabLongTail}{%
\begin{table}[!htbp]
\centering
\caption{Accuracy on long-tailed ImageNet streams with ResNet50 and ViT-B/16.}
\label{tab:long-tail}
\small
\begin{tabular}{lllccccc}
\toprule
Backbone & Stream & Group & CLIP & TDA & OGA & BCA & TTC \\
\midrule
\multirow{4}{*}{ResNet50} & \multirow{2}{*}{$\sim$21K (3:1)} & Overall & 61.20 & 61.82 & 60.23 & 61.27 & \textbf{61.94} \\
 &  & Tail 20\% & 60.81 & 61.55 & 60.49 & 61.18 & \textbf{62.22} \\
\cmidrule(l){2-8}
 & \multirow{2}{*}{$\sim$12K (6:1)} & Overall & 61.26 & \textbf{61.58} & 59.76 & 61.36 & 61.21 \\
 &  & Tail 20\% & 60.32 & 60.29 & 58.19 & 60.80 & \textbf{62.47} \\
\midrule
\multirow{4}{*}{ViT-B/16} & \multirow{2}{*}{$\sim$21K (3:1)} & Overall & 69.66 & 70.12 & 69.54 & 69.76 & \textbf{70.96} \\
 &  & Tail 20\% & 69.66 & 69.97 & 70.12 & 69.87 & \textbf{72.11} \\
\cmidrule(l){2-8}
 & \multirow{2}{*}{$\sim$12K (6:1)} & Overall & 69.74 & \textbf{70.08} & 68.99 & 69.84 & 69.66 \\
 &  & Tail 20\% & 68.92 & 69.19 & 67.49 & 69.17 & \textbf{70.89} \\
\bottomrule
\end{tabular}
\end{table}
}

\usepackage{amsmath,amsfonts,bm,xspace}

\DeclareRobustCommand\onedot{\futurelet\@let@token\@onedot}
\def\@onedot{\ifx\@let@token.\else.\null\fi\xspace}
 
\def\ie{\emph{i.e}.}

\DeclareMathOperator*{\argmax}{arg\,max}
\DeclareMathOperator*{\argmin}{arg\,min}

\DeclareMathAlphabet{\mathsfit}{\encodingdefault}{\sfdefault}{m}{sl}
\SetMathAlphabet{\mathsfit}{bold}{\encodingdefault}{\sfdefault}{bx}{n}

\begin{document}
\maketitle

\begin{abstract}
Vision-language models (VLMs) achieve strong zero-shot transferability but remain vulnerable to target-domain shifts at inference time. Test-time adaptation (TTA) offers a practical remedy, yet most existing VLM-TTA methods follow a prediction-side adaptation paradigm. They use test samples to adjust logits, prototypes, caches, priors, or feature statistics, often incurring additional computational overhead. In this paper, we take a different perspective and reframe VLM-TTA as candidate verification rather than prediction adjustment. We propose \textbf{Test-Time Correction (TTC)}, a hypothesis-based correction framework guided by a simple principle: \textit{hypothesize, reconstruct, correct}. Given a test feature and its top-$k$ candidate labels, TTC treats each candidate label as a hypothesis, reconstructs the feature within the corresponding latent subspace stored in a memory bank, and measures the resulting \textbf{divergence shift}. This shift quantifies how much the candidate subspace and its relations to other candidates change after the hypothetical insertion of the test feature. A correct candidate hypothesis induces only a small shift, whereas an incorrect one perturbs the subspace more strongly. TTC therefore corrects the prediction by selecting the candidate with the minimum aggregated divergence shift. This training-free candidate-verification mechanism avoids iterative optimization and provides a favorable accuracy-efficiency trade-off. Across five TTA settings and 15 benchmark datasets, including zero-shot classification, domain generalization, few-shot classification, base-to-novel generalization, and cross-dataset evaluation, TTC consistently improves accuracy over state-of-the-art VLM-TTA methods while achieving up to $2\times$ speedup, over $3\times$ lower CPU memory usage, and up to $1.4\times$ lower GPU memory usage than the lowest-memory training-free baseline.\footnote{Our code is available at \url{https://github.com/mmai-laboratory/ttc}}
\end{abstract}

\section{Introduction}
\label{sec:introduction}
Vision-language models (VLMs)~\citep{radford2021clip,jia2021align,alayrac2022flamingo,li2021albef,yu2022coca,singh2022flava} have shown strong zero-shot transferability across diverse visual recognition tasks by aligning image and text representations in a shared embedding space. However, they remain vulnerable to target-domain shifts, including changes in style, texture, viewpoint, and data distribution. Test-time adaptation (TTA) has therefore emerged as a practical strategy for improving VLM predictions at inference time using only unlabeled test samples.

Existing VLM-based TTA methods adapt different components of the prediction pipeline. Early approaches~\citep{shu2022tpt,ma2023swapprompt,liu2024dart,yoon2024ctpt,sharifdeen2025otpt,zhang2025scap,zhao2023rlcf,abdul2023align,zhang2024histpt,feng2023difftpt} optimize learnable prompts or feature representations, often using multiple augmented views of each test sample. Recent training-free methods~\citep{karmanov2024tda,fuchs2025oga,zhou2025bca} reduce this overhead through cache retrieval, prototype refinement, or distribution modeling. Despite their differences, most methods share the same principle. They use test samples to adjust logits, prototypes, or feature statistics, mainly adapting on the prediction side.

In this work, we take a different perspective. Rather than using test samples to adjust predictions, we verify whether the test sample is geometrically compatible with each candidate class. Given a test feature and its top-$k$ candidate labels, we treat each candidate as a hypothesis and examine whether its class subspace can accommodate the test feature with minimal perturbation. This reframes TTA as candidate verification rather than prediction adjustment.
To instantiate this idea, we propose \textbf{Test-Time Correction (TTC)}, a hypothesis-based correction framework for VLM-TTA. TTC follows a simple principle: \textit{hypothesize, reconstruct, correct}. It first selects the top-$k$ candidates from the original VLM prediction. For each candidate hypothesis, TTC reconstructs the test feature within the corresponding latent subspace stored in a memory bank and measures the resulting \textbf{divergence shift}. This shift quantifies how much the candidate subspace and its relations to other candidate subspaces change after the hypothetical insertion. 
If the hypothesis is correct, the insertion should induce only a small shift; otherwise, it should perturb the subspace more strongly. TTC then corrects the prediction by selecting the candidate with the minimum aggregated divergence shift.

This candidate-verification view distinguishes TTC from prior VLM-TTA methods. Unlike confidence-based approaches, TTC does not assume that high logits remain reliable under domain shift. It explicitly tests whether each candidate class subspace can accommodate the test feature with minimal structural perturbation.
The divergence-shift criterion is motivated by subspace perturbation analysis~\citep{kato2013perturbation,yu2015useful,davis1963rotation}. Adding an in-class sample should preserve the corresponding class subspace, whereas adding an out-of-class sample should cause a larger perturbation. Since TTC operates only over a small top-$k$ candidate set and avoids iterative optimization or heavy test-time processing, it offers a favorable accuracy-efficiency trade-off, achieving up to $2\times$ speedup and over $3\times$ lower CPU memory usage than prior VLM-TTA methods while retaining a distinctive candidate-verification mechanism.
Across five TTA settings and 15 benchmark datasets, TTC consistently improves accuracy while maintaining competitive inference efficiency and low memory usage. These results show that candidate verification is an effective alternative to conventional prediction-side adaptation.
Our contributions are as follows:
\begin{itemize}
\item We propose \textbf{Test-Time Correction (TTC)}, a hypothesis-based correction framework that reframes VLM-TTA as candidate verification rather than prediction-side adaptation.
\item We introduce \textbf{divergence shift}, a perturbation-based criterion that verifies each candidate label by measuring how stably the test feature can be reconstructed within the corresponding candidate subspace after its hypothetical insertion into that subspace. 
\item We demonstrate that TTC achieves a favorable accuracy-efficiency trade-off, attaining state-of-the-art performance with up to $2\times$ speedup, over $3\times$ lower CPU memory, and up to $1.4\times$ lower GPU memory than the most efficient training-free baselines.
\end{itemize}

\section{Related Work}
\subsection{Test-Time Adaptation in VLMs}
Vision-language models (VLMs) have shown strong generalization, motivating the development of parameter-efficient adaptation methods for downstream datasets, such as prompt learning and feature adaptation~\citep{zhou2022coop, zhou2022cocoop, pratt2023does}.
Beyond offline adaptation, test-time adaptation (TTA) improves VLM predictions using only pre-trained models and unlabeled test data. Early VLM-TTA methods rely on test-time optimization or augmentation. TPT~\citep{shu2022tpt} optimizes learnable prompts using augmented views, while DiffTPT~\citep{feng2023difftpt} improves view diversity with diffusion-generated images.
Later methods refine prediction-side adaptation by calibrating textual features~\citep{yoon2024ctpt}, imposing prompt orthogonality~\citep{sharifdeen2025otpt}, aligning prediction distributions~\citep{abdul2023align}, or updating auxiliary representations~\citep{liu2024dart,ma2023swapprompt,zhang2024histpt,zhang2024dual,zhang2025scap}. However, these methods require iterative optimization, multiple views, or additional computation.

Recent training-free methods reduce this cost by replacing backpropagation with memory-, prototype-, distribution-, or prior-based adjustment. MTA~\citep{zanella2024mta} builds mean-shift prototypes from augmented prediction scores, TDA~\citep{karmanov2024tda} uses positive and negative caches, OGA~\citep{fuchs2025oga} models online Gaussian statistics, and BCA~\citep{zhou2025bca} updates class-level likelihoods or priors through Bayesian class adaptation. Although these methods improve practicality, they still operate as prediction-side adjustments.

TTC takes a different perspective. Rather than primarily adjusting prediction statistics, it verifies candidate labels by reconstructing the test feature within each candidate subspace and measuring the resulting reconstruction-induced divergence shift.

\subsection{Nearest Subspace Reconstruction}
Nearest-subspace methods~\citep{naseem2010linear,yu2024ste,basri2007approximate} measure the proximity of a query to a class-specific subspace rather than to individual data points. A class subspace is constructed from representative features of the same class, and the query is projected onto this subspace to estimate its reconstruction error. Compared with point-wise similarity or nearest-neighbor matching~\citep{arya1998optimal,andoni2018approximate}, which can be sensitive to noise in high-dimensional spaces, subspace representations provide geometric robustness by capturing class-level structure and suppressing local noise. This makes nearest-subspace methods useful for robust pattern recognition~\citep{naseem2010linear,basri2007approximate} and geometric estimation~\citep{yu2024ste}.

TTC is related to nearest-subspace methods, but it uses reconstruction in a different way. Conventional nearest-subspace classifiers assign a query to the class with the smallest absolute reconstruction error~\citep{basri2007approximate,naseem2010linear}. In contrast, TTC uses reconstruction as a perturbation probe for candidate verification. The decision depends not only on the distance between the test sample and a class subspace, but also on how stably the candidate subspace accommodates the sample without disrupting its relations to other candidates. This distinction is central to TTC. An in-class sample should be reconstructed within its corresponding class subspace with only a small perturbation, whereas an out-of-class sample should induce a greater change in subspace relations. This intuition is consistent with subspace perturbation analysis~\citep{davis1963rotation,kato2013perturbation,yu2015useful}.
TTC quantifies this behavior through divergence shift and uses it as a criterion for correcting candidate labels. It thus extends nearest-subspace methods from absolute proximity estimation to candidate verification through reconstruction-induced divergence shift.

\figMain

\section{Method}
\label{sec:method}
We present \textbf{Test-Time Correction (TTC)}, a candidate-verification framework for VLM-TTA, as illustrated in \cref{fig:overview}. Given a test feature, TTC obtains the original VLM logits and selects the top-$k$ candidate labels. Instead of directly trusting the top-$1$ prediction or updating prediction-side statistics, TTC treats each candidate label as a hypothesis and verifies it through subspace stability. For each candidate, TTC hypothetically inserts the test feature into the corresponding class memory, estimates the resulting divergence shift against other candidate subspaces, and selects the candidate with the minimum aggregated verification score. The overall correction rule is
\begin{equation}
\label{eq:correction-rule}
\hat{y} \;=\; \argmin_{y \in \mathcal{C}_k(\mathbf{x})}\, E_y(\mathbf{x}),
\end{equation}
where $\mathcal{C}_k(\mathbf{x})$ is the top-$k$ candidate set for the test feature $\mathbf{x}$, and $E_y(\mathbf{x})$ denotes the aggregated verification score for candidate label $y$, formally defined in Definition~\ref{def:agg-score}.

\subsection{Candidate Hypothesis and Memory Subspace}
\label{sec:method-candidates}

Let $\mathbf{x} \in \mathbb{R}^{d}$ be the visual feature of a test image extracted by a frozen VLM, and let $\mathbf{T} = [\mathbf{t}_1, \ldots, \mathbf{t}_{|\mathcal{Y}|}]^\top \in \mathbb{R}^{|\mathcal{Y}| \times d}$ denote the text embeddings of the class label set $\mathcal{Y}$. The original VLM logits are computed as $\mathbf{z} = \mathbf{T} \mathbf{x} \in \mathbb{R}^{|\mathcal{Y}|}$, and the initial prediction is $\hat{y}_0 \;=\; \argmax_{y \in \mathcal{Y}}\, z_y$.
TTC does not directly accept this $\hat{y}_0$. Instead, it constructs the top-$k$ candidate set
\begin{equation}
\label{eq:topk}
\mathcal{C}_k(\mathbf{x}) \;=\; \mathrm{TopK}_{y \in \mathcal{Y}}(\mathbf{z}),
\end{equation}
and treats each candidate label $y \in \mathcal{C}_k(\mathbf{x})$ as a candidate hypothesis $h_y: \mathbf{x} \mapsto y$. TTC aims to verify which hypothesis $h_y$ is geometrically compatible with the test feature.
To enable this verification, TTC associates each class with a set of representative features stored in a memory bank, and treats the geometric structure induced by these features as the carrier of class identity.
For each class $y \in \mathcal{Y}$, let $\mathbf{B}_y = [\mathbf{b}_{y,1}, \ldots, \mathbf{b}_{y,m_y}]^\top \in \mathbb{R}^{m_y \times d}$ denote the memory matrix storing $m_y \leq m$ feature vectors of class $y$. The associated class memory subspace is $\mathcal{S}_y=\mathrm{span}(\mathbf{B}_y)$.
Verifying a candidate label thus amounts to testing whether $\mathbf{x}$ can be stably reconstructed by the corresponding subspace.

\paragraph{Memory update.}
We maintain $|\mathbf{B}_y| \leq m$ for each class $y$ throughout inference. When a test feature is assigned to class $\hat{y}$ via Eq.~\eqref{eq:correction-rule} and $|\mathbf{B}_{\hat{y}}| < m$, the feature is appended to $\mathbf{B}_{\hat{y}}$. Once the memory is full, $|\mathbf{B}_{\hat{y}}| = m$, the entry with the lowest associated confidence is evicted to make room for the new feature, ensuring that $\mathbf{B}_{\hat{y}}$ remains a high-confidence representation of class $\hat{y}$. In label-free settings, the memory is initialized from confident model predictions on test samples. When labeled training samples are available, their features are stored as fixed memory entries.

\subsection{Hypothetical Insertion and Divergence Shift}
\label{sec:method-shifts}

For each candidate hypothesis $h_y$, TTC hypothetically inserts the test feature into the candidate class memory and forms an updated subspace
\begin{equation}
\label{eq:insertion}
\widetilde{\mathbf{B}}_y \;=\; [\mathbf{B}_y;\, \mathbf{x}], \qquad \widetilde{\mathcal{S}}_y \;=\; \mathrm{span}(\widetilde{\mathbf{B}}_y).
\end{equation}
If $h_y$ is correct, $\mathbf{x}$ should already be well explained by $\mathcal{S}_y$, and the updated subspace $\widetilde{\mathcal{S}}_y$ should preserve the original reconstruction relations of $\mathcal{S}_y$. If $h_y$ is incorrect, inserting $\mathbf{x}$ into $\mathcal{S}_y$ introduces a larger out-of-subspace component and disrupts these relations more strongly.
This intuition follows the classical subspace perturbation view~\citep{davis1963rotation, yu2015useful,kato2013perturbation}. To turn this single-subspace perturbation into a verification signal, TTC compares subspace relations \emph{before} and \emph{after} insertion.

\begin{definition}[Pairwise Divergence Shift]
\label{def:pairwise-shift}
Let $d_{\mathcal{S}}(\cdot, \cdot)$ denote a reconstruction-based divergence between subspaces. Given a test feature $\mathbf{x}$ and two candidate labels $y, r \in \mathcal{C}_k(\mathbf{x})$ with $y \neq r$, the \emph{pairwise divergence shift} from $y$ to $r$ induced by inserting $\mathbf{x}$ into the memory of class $y$ is
\begin{equation}
\label{eq:pairwise-shift}
\delta_{y \to r}(\mathbf{x}) \;=\; d_{\mathcal{S}}(\mathcal{S}_y,\, \mathcal{S}_r) \;-\; d_{\mathcal{S}}(\widetilde{\mathcal{S}}_y,\, \mathcal{S}_r).
\end{equation}
The reverse shift $\delta_{r \to y}(\mathbf{x})$ is defined analogously by inserting $\mathbf{x}$ into $\mathbf{B}_r$ instead.
\end{definition}

The two directions provide complementary evidence. If $\mathbf{x}$ truly belongs to candidate $y$, then $\delta_{y \to r}(\mathbf{x})$ should remain small because inserting $\mathbf{x}$ into $\mathbf{B}_y$ barely perturbs $\mathcal{S}_y$. At the same time, $\delta_{r \to y}(\mathbf{x})$ should be large because inserting $\mathbf{x}$ into a competing subspace $\mathcal{S}_r$ introduces a substantial orthogonal residual. Combining both directions yields a more discriminative signal than either alone.

\subsection{Aggregated Verification Score}
\label{sec:method-aggregation}
We now combine absolute reconstruction compatibility and pairwise divergence shifts into a single score per candidate. Because the score involves only the $k$ candidate memories, each test sample requires $k(k-1)$ pairwise shift evaluations and no gradient updates.

\begin{definition}[Aggregated Verification Score]
\label{def:agg-score}
For each candidate label $y \in \mathcal{C}_k(\mathbf{x})$, the \emph{aggregated verification score} is
\begin{equation}
E_y(\mathbf{x}) \;=\; \underbrace{d_{\mathbf{x}}(\mathcal{S}_y,\, \mathbf{x})}_{\text{absolute compatibility}} \;+\; \underbrace{\sum_{r \in \mathcal{C}_k(\mathbf{x}) \setminus \{y\}} \bigl[ \delta_{y \to r}(\mathbf{x}) - \delta_{r \to y}(\mathbf{x}) \bigr]}_{\text{relative stability}},
\label{eq:agg-score}
\end{equation}
where $d_{\mathbf{x}}(\mathcal{S}_y, \mathbf{x})$ measures the reconstruction discrepancy between $\mathbf{x}$ and $\mathcal{S}_y$. The TTC prediction is then $\hat{y} = \argmin_{y \in \mathcal{C}_k(\mathbf{x})} E_y(\mathbf{x})$.
\end{definition}

The first term anchors the score to the absolute compatibility between $\mathbf{x}$ and the candidate subspace, ensuring that geometrically distant candidates are penalized even when their pairwise shifts are small. The second term contrasts the stability of candidate $y$ with competing hypotheses $r$. Specifically, subtracting $\delta_{r \to y}(\mathbf{x})$ rewards candidates for which insertion into competing subspaces causes large perturbations. A correct candidate should achieve both a small absolute discrepancy and a small pairwise divergence shift, resulting in a low $E_y(\mathbf{x})$.

\subsection{Subspace Divergence Metric}
\label{sec:subspace-divergence-metric}

TTC can be instantiated with different subspace divergence choices. We adopt a nearest-subspace reconstruction error, which directly measures how well one class reconstructs features from another. For a class memory $\mathbf{B}_y$, we define the ridge-regularized projection matrix as
\begin{equation}
\mathbf{P}_y=\mathbf{B}_y^\top \bigl( \mathbf{B}_y \mathbf{B}_y^\top + \lambda \mathbf{I} \bigr)^{-1} \mathbf{B}_y,
\end{equation}
where $\lambda > 0$ is a regularization coefficient. The divergence from $\mathcal{S}_a$ to $\mathcal{S}_b$ is computed as the average row-wise reconstruction error of features in $\mathbf{B}_b$ projected onto $\mathcal{S}_a$ as
\begin{equation}
d_{\mathcal{S}}(\mathcal{S}_a,\, \mathcal{S}_b) \;=\; \frac{1}{m_b} \,\bigl\| \mathbf{B}_b \mathbf{P}_a - \mathbf{B}_b \bigr\|_{2,1},
\label{eq:nse-subspace}
\end{equation}
where $\| \cdot \|_{2,1}$ denotes the sum of row-wise Euclidean norms.
$d_\mathcal{S}$ depends on the memories rather than only on their spans; we keep the subspace notation for readability.
The feature-to-subspace discrepancy in Eq.~\eqref{eq:agg-score} is defined analogously
\begin{equation}
d_{\mathbf{x}}(\mathcal{S}_y,\, \mathbf{x}) \;=\; \bigl\| \mathbf{x}^\top \mathbf{P}_y - \mathbf{x}^\top \bigr\|_2.
\label{eq:nse-feature}
\end{equation}
With this metric, the effect of a hypothetical insertion admits a closed form, which we derive and discuss in \cref{app:perturbation}.

\section{Experiments}
\label{sec:experiment}
The experiments evaluate whether reconstruction-induced divergence shifts provide an effective and efficient candidate-verification signal across diverse distribution shifts and transfer scenarios. In all settings, TTC keeps the VLM frozen and performs correction solely through candidate verification. We also compare computational efficiency against recent VLM-TTA methods and conduct ablations on the divergence metric, verification score, top-$k$, and memory design.

\subsection{Experimental Setup}
To rigorously evaluate our method across diverse transfer learning scenarios, we use a comprehensive suite of five datasets~\citep{ImageNet,hendrycks2021imagenetA,hendrycks2021imagenetR,wang2019imagenetSketch,recht2019imagenetV2} for domain generalization. For zero-shot classification, few-shot classification, base-to-novel generalization, and cross-dataset evaluation, we use 11 downstream datasets~\citep{li2022caltech,dtd,nilsback2008flowers102,ImageNet,soomro2012ucf101,fgvc,helber2018eurosat,food101,xiao2010sun397,stanfordcars,oxfordpetiii} with the ResNet50 and ViT-B/16 backbones of CLIP~\citep{radford2021clip}. When using a few training samples, namely in few-shot, base-to-novel, and cross-dataset settings, we strictly follow the standard protocols established in prior work~\citep{zhou2022coop,shu2022tpt}.
We repeat every experiment with three random seeds, each of which randomly permutes the entire test stream, and report the average.
We use the top-$k=3$ candidate labels and fix the maximum memory size to $m=16$. The memory bank initialization protocol depends on the availability of training data. Specifically, when no training samples are available (\ie, zero-shot and domain generalization), the memory bank is initialized using the model's original predictions on test samples. Conversely, when training data is available (\ie, few-shot, base-to-novel, and cross-dataset), the memory bank is initialized using the corresponding training samples. \Cref{app:experimental-details} provides further details.

\subsection{Experimental Results}
\tabZeroShot
\paragraph{Zero-shot classification.}
\label{sec:zero-shot}
We evaluate whether TTC improves the initial zero-shot predictions of VLMs.
As shown in \cref{tab:zeroshot}, TTC consistently outperforms state-of-the-art VLM-TTA methods~\citep{shu2022tpt,feng2023difftpt,yoon2024ctpt,sharifdeen2025otpt,sui2025tps,farina2024zero,karmanov2024tda,fuchs2025oga,zhang2024dual,zhou2025bca} across both ResNet50 and ViT-B/16 backbones. Among training-free methods, TTC achieves gains of 2.5 percentage points (pp) and 2.4\,pp, respectively. These results demonstrate that candidate verification provides a discriminative correction signal beyond prediction adjustments based on caches, prototypes, or priors.

\tabDomainGen
\paragraph{Domain generalization.}
We further evaluate TTC under natural distribution shifts. \Cref{tab:domaingen} reports both zero-augmentation and augmentation-enhanced variants. Although prior methods~\citep{shu2022tpt,karmanov2024tda,zhang2024dual} often use 63 augmented views, TTC uses fewer augmentations while outperforming other training-free methods by 1.0\,pp with ResNet50 and 1.3\,pp with ViT-B/16. Under the zero-augmentation protocol, TTC surpasses existing methods by 2.3\,pp and 1.9\,pp with ResNet50 and ViT-B/16, respectively.

\figFewshot
\paragraph{Few-shot classification.}
\label{sec:few-shot}
We apply TTC on top of CoOp~\citep{zhou2022coop}, which is trained with different numbers of shots per class, using the ViT-B/16 CLIP backbone. 
\cref{fig:fewshot} shows that TTC consistently improves performance across training shot numbers. 
These gains show that TTC complements few-shot prompt learning. Rather than modifying the learned prompt, TTC corrects each test-time prediction by verifying which candidate subspace most stably accommodates the test feature. 
This suggests that TTC can provide additional robustness even when the model has already been adapted with a small amount of labeled data. For detailed results, refer to \cref{app:few-shot}.

\tabBaseToNovelAndCrossDataset
\paragraph{Base-to-novel generalization.}
\label{sec:base2novel}
Base-to-novel generalization evaluates whether a method can improve novel-class recognition without compromising the knowledge learned from base classes. \cref{tab:base2novel} reports the harmonic mean of base and novel accuracies across 11 datasets. TTC achieves the best average harmonic mean, improving over the state-of-the-art by 0.9\,pp. This result indicates that candidate verification can mitigate prediction bias in prompt-tuned models while preserving base-class performance. Base- and novel-class results are provided in \cref{app:base-to-novel}.

\paragraph{Cross-dataset evaluation.}
We evaluate cross-dataset transfer, where CoOp is trained on ImageNet and tested on other downstream datasets. 
As shown in \cref{tab:crossdataset}, TTC improves the best prior VLM-TTA baseline by approximately 0.7\,pp on average.
These performance gains show that TTC is not tied to a specific target dataset and remains effective under cross-dataset distribution shifts.

\tabResources
\subsection{Efficiency Comparison}
We analyze whether the proposed candidate-verification mechanism can be performed efficiently at test time. \cref{tab:resources} compares TTC with recent VLM-TTA methods on ImageNet zero-shot classification. To ensure a fair comparison of computational resources, we pre-extract test features and do not use augmentation for all methods, thereby isolating the resources required by each adaptation algorithm. TTC achieves the highest accuracy, uses the least memory, and delivers the highest throughput among compared methods. With ViT-B/16, TTC achieves $2\times$ the throughput of BCA while reducing CPU memory usage by $3\times$. With ResNet50, TTC improves both accuracy and throughput over BCA while requiring substantially less memory. Across both backbones, TTC uses $1.3$--$1.4\times$ less GPU memory than the lowest-memory training-free baseline.

\subsection{Analysis}

\tabHitRate
\paragraph{Top-$k$ hit rate and decomposition.}
TTC re-ranks only the top-$k$ candidates, so its accuracy on each dataset decomposes as
$\mathrm{Acc} = \mathrm{Hit@}k \times P(\mathrm{correct}\mid\mathrm{GT}\in\text{top-}k),$
where $\mathrm{Hit@}k$ is the fraction of samples whose ground truth lies in the top-$k$ set, and the second factor, $A_{\mathrm{in}}$, measures candidate-verification accuracy. As shown in \cref{tab:topk-summary}, in most settings, the top-$3$ set contains the true label for 80--94\% of samples, while the initial prediction misses this label on 13--22\% of all samples, leaving substantial room for correction by TTC. Within this space, TTC improves $A_{\mathrm{in}}$ over the initial prediction by an average of 3.0\,pp, compared with 1.3\,pp for the strongest baseline in each setting. By contrast, $A_{\mathrm{out}}$ shows that correctly predicting samples outside the top-$3$ is difficult.

\tabTaxonomy
\paragraph{Success and failure study.}
We partition predictions within the hit set into three cases in \cref{tab:taxonomy}.
Overturning a correct initial prediction is rarest at 3.3\% and occurs mainly with nearly duplicate labels such as \textit{sunglass} and \textit{sunglasses}. Both failure cases have verification margins 3.4--6.5$\times$ smaller than those of successes, indicating that failures concentrate where competing candidates are nearly tied. These candidates typically represent semantically similar classes such as \textit{ear of corn} and \textit{corn}.
The detailed results are provided in \cref{app:success-and-failure}.

\subsection{Ablation Study}
We conduct ablation studies to analyze why TTC works in the zero-shot classification setting. Specifically, we evaluate different divergence metrics, the contribution of each term in the aggregated verification score, and the sensitivity of TTC to top-$k$, memory size, and memory quality. In these studies, the average denotes the mean accuracy across 11 datasets, and detailed results are provided in \cref{app:memory-robustness,app:divergence-metric-functions-aggregated-verification-score}.

\begin{figure}[!t]
\begin{minipage}{0.48\linewidth}
\tabMetricAblationZS
\end{minipage}
\hfill
\begin{minipage}{0.48\linewidth}
\tabFunctionAblationZS    
\end{minipage}
\end{figure}

\paragraph{Divergence metrics.}
We compare three divergence metrics: $l_2$ distance, cosine similarity (Cos. Sim.), and nearest-subspace reconstruction error (NSE). When cosine similarity is used, we negate the similarity value so that lower values consistently indicate better compatibility in the aggregated verification score. As shown in \cref{tab:metric-ablation-zs}, NSE achieves the best average accuracy for both ResNet50 and ViT-B/16. This confirms that reconstruction-based subspace divergence is well aligned with the candidate-verification principle of TTC.

\paragraph{Aggregated verification score.}
\label{sec:ablation-aggregated-divergence-shifts}
We analyze the contribution of each component in $E_y(\mathbf{x})$. \cref{tab:function-ablation-zs} compares the anchor reconstruction term $d_{\mathbf{x}}(\mathcal{S}_y,\mathbf{x})$, the forward divergence shift $\delta_{y \to r}(\mathbf{x})$, and the reverse divergence shift $\delta_{r \to y}(\mathbf{x})$. Each component improves performance when used individually, while combining all three achieves the best accuracy. This result demonstrates that the three terms, absolute compatibility, candidate stability, and competing-hypothesis perturbation, provide complementary evidence for candidate verification.

\paragraph{Top-\textit{k}.}
\cref{fig:topk} illustrates the effect of varying the number of candidate labels. Increasing $k$ gives TTC more candidate hypotheses to verify, but also increases computation.  Accuracy improves until $k=3$, after which the gains become marginal. 
\figAblation

\paragraph{Memory size.}
\cref{fig:memsize-inet,fig:memsize} analyze the effect of the maximum per-class memory size $m$. Since some datasets contain fewer than 16 samples per class, \cref{fig:memsize-inet} reports the accuracy on ImageNet, while \cref{fig:memsize} reports the average accuracy across 11 datasets.
More memory provides a richer estimate of each class subspace but also increases reconstruction cost. TTC achieves substantial gains even with a small memory, and its performance saturates around $m=16$.

\paragraph{Memory quality.}
We examine how noisy memory initialization affects TTC. This analysis matters in label-free settings because the memory is initialized with pseudo-labels produced by the frozen VLM, which may include misclassified features. TTC is robust to such noise because it does not make decisions based on individual memory entries. Instead, it verifies candidates by aggregating reconstruction-induced divergence shifts over class subspaces.

\cref{fig:memacc} shows that TTC remains effective even with imperfect memory initialization. The dashed line denotes a static memory that is not updated after initialization and is intentionally kept fixed to preserve its initial purity throughout evaluation. In contrast, the solid line represents the setting in which memory is continuously updated during inference. When initial memory purity is low, online memory updates outperform static memory, suggesting that corrected test samples improve the memory structure. These results indicate that TTC does not require perfectly clean memory. Instead, a coarse and gradually refined class subspace is sufficient for candidate verification.

\section{Conclusion}
\label{sec:conclusion}
This paper introduced \textbf{Test-Time Correction (TTC)}, a candidate-verification framework for VLM test-time adaptation. Rather than adapting prediction-side statistics, TTC treats top-$k$ candidate labels as hypotheses and verifies them by measuring reconstruction-induced divergence shifts after hypothetical insertion into class memories. This reframes test-time adaptation as a geometric verification problem, where a correct candidate should accommodate the test feature with minimal perturbation, whereas an incorrect candidate should more strongly disturb candidate subspace relations.
TTC performs correction without updating model parameters or prompts and operates only over a small candidate set with compact class memories. Across zero-shot classification, domain generalization, few-shot classification, base-to-novel generalization, and cross-dataset evaluation, TTC consistently improves VLM predictions while maintaining strong computational efficiency. These results demonstrate that candidate verification provides an effective and practical alternative to conventional prediction-side adaptation.

TTC relies on the original VLM logits to construct its top-$k$ candidate set and therefore cannot recover the correct label when the true class is excluded. Future work could explore adaptive candidate expansion, uncertainty-aware selection, or memory-guided retrieval beyond the initial predictions. Moreover, for unseen classes or short streams with sparse or empty candidate memories, TTC defaults to the initial prediction, limiting its potential gains.

\section*{Acknowledgments}
This research was supported by the National Research Foundation of Korea (NRF), Electronics and Telecommunications Research Institute(ETRI), and Institute of Information \& Communications
Technology Planning \& Evaluation (IITP), funded by the Korean government [26CS1100, Development of Proprietary Physical AI-based Small-scale Computers and Integrated Soft Suits], and the Korean government(MSIT) (RS-2026-25518808, RS-2026-25617480, and IITP-2026-RS-2023-00255968).

\bibliographystyle{plainnat}
\bibliography{main}

\newpage
\clearpage
\appendix
\crefalias{section}{appendix}
\crefalias{subsection}{appendix}

\section{Perturbation View of Divergence Shift}
\label{app:perturbation}
We provide a perturbation-based motivation for the divergence-shift criterion used in TTC. 
This section does not establish a strict classification guarantee; it explains why hypothetical insertion provides a meaningful candidate-verification signal. 
The argument proceeds in three steps: 
(i) a geometric intuition of how inserting a test feature perturbs a candidate subspace, 
(ii) a closed-form perturbation view relating this change to the residual of the test feature, and 
(iii) an explanation of how reconstruction-induced divergence shift operationalizes this stability principle.

\subsection{Intuition}
For each class $y$, let $\mathbf{B}_y \in \mathbb{R}^{m_y \times d}$ denote the class memory matrix and let $\mathcal{S}_y = \mathrm{span}(\mathbf{B}_y)$ be the class subspace induced by the memory features. 
Given a test feature $\mathbf{x}$, TTC verifies a candidate hypothesis $h_y : \mathbf{x} \mapsto y$ by hypothetically inserting $\mathbf{x}$ into $\mathbf{B}_y$. 
This yields the updated memory and the updated subspace as 
\begin{equation}
\widetilde{\mathbf{B}}_y = [\mathbf{B}_y;\mathbf{x}], \qquad \widetilde{\mathcal{S}}_y = \mathrm{span}(\widetilde{\mathbf{B}}_y).    
\end{equation}

The change from $\mathcal{S}_y$ to $\widetilde{\mathcal{S}}_y$ depends on how much of $\mathbf{x}$ lies outside the original subspace $\mathcal{S}_y$. 
Let $\mathrm{Proj}_{\mathcal{S}_y}(\mathbf{x})$ denote the projection of $\mathbf{x}$ onto $\mathcal{S}_y$, and define the residual component as
\begin{equation}
\mathbf{x}_{\perp y} = \mathbf{x} - \mathrm{Proj}_{\mathcal{S}_y}(\mathbf{x}).
\end{equation}
If $\mathbf{x}$ truly belongs to class $y$, then it should be geometrically compatible with $\mathcal{S}_y$, and the residual norm $\|\mathbf{x}_{\perp y}\|$ is expected to be small. 
In this case, inserting $\mathbf{x}$ into $\mathbf{B}_y$ only mildly changes the induced subspace. 
In contrast, if $\mathbf{x}$ belongs to another class, the residual component with respect to $\mathcal{S}_y$ is expected to be larger, and the updated subspace $\widetilde{\mathcal{S}}_y$ deviates more strongly from the original subspace.

\subsection{Subspace-Perturbation View}
This intuition aligns with classical subspace perturbation results~\citep{davis1963rotation, yu2015useful}, which bound subspace change by the perturbation size relative to a spectral gap. For exact spans, however, the principal angles between $\mathcal{S}_y$ and $\widetilde{\mathcal{S}}_y$ are not informative: $\mathcal{S}_y \subseteq \widetilde{\mathcal{S}}_y$ holds by construction, and the orthogonal projector onto the span changes by a unit-norm rank-one projector whenever $\mathbf{x} \notin \mathcal{S}_y$, regardless of how small $\|\mathbf{x}_{\perp y}\|$ is.

The ridge-regularized projection matrix $\mathbf{P}_y$ of \cref{sec:subspace-divergence-metric} removes this discontinuity. Let $\mathbf{r}_y = (\mathbf{I} - \mathbf{P}_y)\mathbf{x}$ be the ridge residual of $\mathbf{x}$, so that $\|\mathbf{r}_y\|_2 = d_{\mathbf{x}}(\mathcal{S}_y, \mathbf{x})$, and let $\widetilde{\mathbf{P}}_y$ be the projection matrix of $\widetilde{\mathbf{B}}_y$. By the Sherman--Morrison formula, the hypothetical insertion changes the projection matrix by a rank-one term along the residual,
\begin{equation}
\widetilde{\mathbf{P}}_y - \mathbf{P}_y \;=\; \frac{\mathbf{r}_y \mathbf{r}_y^\top}{\lambda + \mathbf{x}^\top \mathbf{r}_y},
\label{eq:rank-one}
\end{equation}
from which
\begin{equation}
\bigl\| \widetilde{\mathbf{P}}_y - \mathbf{P}_y \bigr\|_2 \;=\; \frac{\|\mathbf{r}_y\|_2^2}{\lambda + \mathbf{x}^\top \mathbf{r}_y} \;\le\; \min\Bigl\{1,\; \tfrac{1}{\lambda}\, d_{\mathbf{x}}(\mathcal{S}_y, \mathbf{x})^2 \Bigr\}.
\end{equation}
Thus, an in-class insertion with a small residual induces only a small change, whereas an out-of-class insertion tends to perturb the candidate subspace more strongly. As $\lambda \to 0$, $\mathbf{r}_y$ approaches $\mathbf{x}_{\perp y}$, and the change approaches the unit-norm projector above. The ridge term is therefore essential for a graded response.

This view also explains the special case in which $\mathbf{x}$ already lies in $\mathcal{S}_y$. 
If $\mathbf{x} \in \mathcal{S}_y$, inserting $\mathbf{x}$ does not enlarge the span of the memory matrix, so $\widetilde{\mathcal{S}}_y = \mathcal{S}_y$. 
The ridge residual is then at most $\lambda / (\sigma_{\min}^2(\mathbf{B}_y) + \lambda) \cdot \|\mathbf{x}\|_2$, where $\sigma_{\min}(\mathbf{B}_y)$ is the smallest nonzero singular value of $\mathbf{B}_y$, so the induced change is small whenever the class memory is well conditioned.

\subsection{Connection to Divergence Shift}
TTC does not directly use the principal-angle deviation as a verification score. Instead, it uses reconstruction-induced divergence shift as an operational proxy for subspace stability. For candidate labels $y$ and $r$, the divergence shift is defined as
\begin{equation}
\delta_{y \rightarrow r}(\mathbf{x}) = d_{\mathcal{S}}(\mathcal{S}_y,\mathcal{S}_r) - d_{\mathcal{S}}(\widetilde{\mathcal{S}}_y,\mathcal{S}_r).    
\end{equation}
This quantity compares the reconstruction-based relation between candidate $y$ and reference candidate $r$ before and after inserting $\mathbf{x}$ into the memory of class $y$.

If $\mathbf{x}$ is compatible with class $y$, the hypothetical insertion only mildly changes $\mathcal{S}_y$. 
Consequently, the reconstruction relation between $\mathcal{S}_y$ and other candidate subspaces is expected to remain stable, producing a small divergence shift. 
If $\mathbf{x}$ is incompatible with class $y$, the insertion introduces a larger out-of-subspace component and changes the induced subspace more strongly. 
This, in turn, is likely to alter the reconstruction relation to other candidate subspaces, producing a larger divergence shift. 
Applying the closed form to each row of $\mathbf{B}_r$ makes this precise:
\begin{equation}
\bigl| \delta_{y \to r}(\mathbf{x}) \bigr| \;\le\; \frac{d_{\mathbf{x}}(\mathcal{S}_y, \mathbf{x})}{\lambda + \mathbf{x}^\top \mathbf{r}_y} \cdot \frac{1}{m_r} \sum_{j=1}^{m_r} \bigl| \mathbf{r}_y^\top \mathbf{b}_{r,j} \bigr|,
\end{equation}
so a large shift requires a residual that is both sizable and aligned with the memory of class $r$.

Divergence shift captures the subspace-stability intuition through TTC's reconstruction operator. TTC measures how hypothetical insertion changes reconstruction behavior among candidate subspaces without explicitly computing principal angles or orthonormal bases. The criterion thus aligns directly with the implemented nearest-subspace reconstruction mechanism.

\subsection{Interpretation}

The perturbation view provides a geometric explanation for TTC's candidate-verification principle. A correct candidate should accommodate the test feature with a small residual-induced perturbation, while an incorrect candidate should induce a larger change in its subspace and reconstruction relations to other candidates. Selecting the candidate with the smallest aggregated divergence shift therefore favors the hypothesis whose class subspace most stably accommodates the test feature.

This interpretation also distinguishes TTC from conventional nearest-subspace classification. A nearest-subspace classifier uses reconstruction error as an absolute measure of proximity between a query and a class subspace. In contrast, TTC uses reconstruction as a perturbation probe to assess how hypothetical insertion of the test feature changes the candidate subspace and its relations to competing candidate subspaces. This perturbation view underlies TTC's candidate-verification mechanism.

\section{Experimental Details}
\label{app:experimental-details}
For performance evaluation on the domain generalization task, we utilize ImageNet~\citep{ImageNet}, ImageNet-A~\citep{hendrycks2021imagenetA}, ImageNet-R~\citep{hendrycks2021imagenetR}, ImageNet-Sketch~\citep{wang2019imagenetSketch}, and ImageNet-V2~\citep{recht2019imagenetV2}. Furthermore, for zero-shot classification, few-shot classification, base-to-novel generalization, and cross-dataset evaluation tasks, we utilize 11 downstream datasets: Caltech~\citep{li2022caltech}, Describable Textures Dataset (DTD)~\citep{dtd}, 102 Category Flower Dataset (Flower)~\citep{nilsback2008flowers102}, ImageNet (INet)~\citep{ImageNet}, UCF101 (UCF)~\citep{soomro2012ucf101}, FGVC-Aircraft Benchmark (Aircraft)~\citep{fgvc}, EuroSAT (ESAT)~\citep{helber2018eurosat}, Food101 (Food)~\citep{food101}, SUN397 (SUN)~\citep{xiao2010sun397}, StanfordCars (Cars)~\citep{stanfordcars}, and Oxford-IIIT Pet (Pets)~\citep{oxfordpetiii}. \cref{tab:dataset-info} shows the number of samples and classes used in experiments.
\tabDatasetInfo

All experiments are conducted using an \textit{Intel(R) Xeon(R) Platinum 8270CL CPU @ 2.70GHz} and an \textit{NVIDIA RTX A5000} GPU with PyTorch 2.6.0, CUDA 12.4, and Python 3.12. To thoroughly demonstrate the effectiveness of our method across diverse tasks, we conduct experiments using CLIP~\citep{radford2021clip}.
We categorize the experimental tasks by their data requirements. For domain generalization and zero-shot classification, no training samples are used. For few-shot classification, we evaluate performance under multiple data regimes, using 1, 2, 4, 8, and 16 training samples per class. For base-to-novel generalization and cross-dataset evaluation, we use a fixed setting of 16 training samples per class.
When using few-shot training samples, we strictly follow the standard protocol adopted in prior work~\citep{zhou2022coop,shu2022tpt}. In this setting, we first train prompts on few-shot samples, then directly evaluate the resulting tuned prompts on classification tasks.
For consistency and fair comparison, we use fixed hyperparameters across all experiments: top-$k=3$ prediction candidates and a memory size of $m=16$. All reported results are averaged over three random seeds, and the test sample order is always shuffled. Baseline results are taken from the original papers where available. 
The baselines that we run use their official implementations and published configurations with the same backbones, datasets, and splits as TTC.

\section{Top-$k$ Candidate Set Hit Rate and Dynamic Top-$k$}
\label{app:top-k-hit-rate}
\Cref{tab:topk-per-dataset} shows that dataset-level conditional accuracy of TTC is higher than the initial top-1 candidate on every dataset except ImageNet-A with ResNet50.
\tabHitRatePerDataset

\paragraph{Dynamic top-$k$.} We examine an entropy-guided dynamic $k$. For each sample, we set $k$ by linear interpolation between $k_{\min}=3$ and $k_{\max}$ based on the Shannon entropy of the softmax distribution. Confident distributions use $k$ near $k_{\min}$, while uncertain distributions use $k$ closer to $k_{\max}$. As shown in \cref{tab:dynamic-k}, this approach yields a small, consistent accuracy gain of 0.18--0.24\,pp over fixed $k=3$, at the cost of 20--43\% lower throughput. Fixed $k=3$ therefore remains preferable when throughput is the priority, while dynamic $k$ offers a modest accuracy gain when the additional cost is acceptable.
\tabDynamicK

\section{Efficiency Analysis}
\label{app:efficiency-analysis}

\paragraph{Verification cost.} Each test sample accesses only its $k=3$ candidate memories with 48 vectors. The per-sample verification cost is $O(k^2m^2d)$, about 5--6\,MFLOPs for ViT-B/16 ($d=512$), three orders of magnitude below one encoder forward pass ($\sim$17\,GFLOPs), and independent of $|\mathcal{Y}|$; only the base logit computation scales with the number of classes. The ridge projector inverts only the $m \times m$ Gram matrix, which is cached per class and refreshed only when the class memory changes, and a hypothetical insertion of $\mathbf{x}$ uses a cached block-inverse update at $O(m^2+md)$, with no new inversion per candidate. \Cref{tab:metric-cost} changes only the divergence metric. Cosine similarity yields only a 1.7$\times$ adaptation-stage speedup at the cost of 0.9--1.2\,pp on ImageNet, and TTC with NSE remains the fastest method in \cref{tab:resources}. The cached projection is therefore not the practical bottleneck.
\tabMetricCost

\paragraph{Full-pipeline efficiency.} In \cref{tab:resources}, all methods, including TDA, process the same single-view, pre-extracted feature stream to isolate adaptation cost under identical inputs. We estimate full-pipeline time on the 50,000 ImageNet test images by adding encoding time, based on measured encoder throughput, to adaptation time. Encoder throughput is 2,136\,img/s for ResNet50 and 1,160\,img/s for ViT-B/16. With 63 augmented views and the original view, TDA takes 1,804\,s and 3,031\,s, respectively, whereas TTC takes 69\,s and 68\,s. TTC is thus approximately $26\times$ and $45\times$ faster than the native protocol of TDA.

\paragraph{Memory footprint.} For ViT-B/16 with $d=512$, the complete ImageNet memory bank requires $|\mathcal{Y}| \times m \times d \times 4$\,B $= 1000 \times 16 \times 512 \times 4$\,B, or approximately 32.8\,MB, plus approximately 1\,MB for the cached $16 \times 16$ Gram inverses. The persistent memory bank and cached inverses reside in CPU memory, while temporary computation is included in the peak GPU memory reported in \cref{tab:resources}. Storage scales linearly with the number of allocated class memories. For 10K open-vocabulary classes, approximately 340\,MB of CPU memory is required, which can be reduced with a smaller $m$ or lower-precision storage. Reducing $m$ from 16 to 8 halves the memory bank size at an accuracy cost of 0.8\,pp for ResNet50 and 0.7\,pp for ViT-B/16 on zero-shot classification.

\section{Memory Robustness}
\label{app:memory-robustness}

\paragraph{Recovery from contaminated memory.} \Cref{fig:memacc} reports final accuracy under noisy memory initialization but does not show whether memory quality recovers during the stream. We therefore conduct a controlled stress test with initial memory purity ranging from 0\% to 30\%. Memory purity is the fraction of stored features whose ground-truth class matches the class memory they are stored in. As in \cref{fig:memacc}, each class memory is initialized with $m=16$ features from the labeled 16-shot training split. The stated fraction comes from the memory's own class, while the remainder comes from other classes sampled uniformly. We record memory purity at 10\% intervals along the stream and macro-average the results across the 11 zero-shot datasets for ResNet50 and ViT-B/16, as shown in \cref{fig:memory-recovery}.
Even when all initial entries are incorrect, memory purity increases steadily from 0\% to 66\% for ResNet50 and 73\% for ViT-B/16. TTC does not collapse, retaining accuracies of 60.00\% and 68.36\%, respectively. Final accuracy increases monotonically with initial purity.
Because the controlled memories are initialized with labeled training features and their incorrect entries come from uniformly random classes, this setting isolates the effect of purity and is not directly comparable to the standard initialization, whose measured initial memory purity is 70.00\%/75.52\%.
\figMemoryRecovery

Three mechanisms reduce sensitivity to contaminated entries. First, confidence-filtered initialization fills each class memory with high-confidence predictions, yielding initial purities of 70.00\% and 75.52\%. Second, within-class eviction replaces the lowest-confidence entry, removing the model's least reliable entries, consistent with the observed purity recovery. Third, ridge-regularized aggregation uses all $m=16$ entries per class, so no decision depends on a single stored feature.

\paragraph{Class-imbalanced test streams.} Memory is partitioned by class, with eviction restricted to each class, so samples assigned to frequent classes cannot displace rare-class entries. Imbalance can still slow the filling of rare-class memories. We evaluate this effect on long-tailed ImageNet streams sampled without replacement according to a Zipf distribution, yielding head:tail frequency ratios of approximately 3:1 and 6:1.
We use ResNet50 and ViT-B/16 with three seeds per ratio, define the tail group as the bottom 20\% of classes by realized frequency, and evaluate all training-free baselines, as reported in \cref{tab:long-tail}.
With ViT-B/16, TTC achieves the largest tail-class gains, improving over CLIP by 2.45 and 1.97\,pp, respectively, compared with at most 0.46\,pp for any baseline. OGA loses 1.43\,pp under the 6:1 imbalance. This pattern holds with ResNet50, with gains of 1.41 and 2.15\,pp. Overall ViT-B/16 accuracy exceeds CLIP by 1.30\,pp on the 3:1 stream but falls below it by 0.08\,pp on the 6:1 stream, where performance on the remaining classes offsets the tail-class gain.
\tabLongTail

\section{Additional Experimental Results}
\label{app:additional-results}
This section provides detailed results that complement \cref{sec:experiment}. We report the per-dataset few-shot results, zero-shot classification and domain generalization with additional CLIP backbones, the base- and novel-class accuracies of base-to-novel generalization, the per-setting success and failure analysis, and the per-dataset ablations of the divergence metric and the aggregated verification score. Across these results, the gains of TTC are not confined to a particular backbone, dataset, or evaluation split.

\subsection{Few-shot Classification}
\label{app:few-shot}
\Cref{fig:fewshot-full} presents additional few-shot classification results on 11 downstream datasets. TTC outperforms state-of-the-art approaches on most datasets, and its advantage persists with more shots.
\figFewshotFull

\subsection{Additional Backbones}
\label{app:additional-backbones}
We also experiment with lighter and larger backbones, ViT-B/32 and ViT-L/14. \Cref{tab:zeroshot-backbones} reports the zero-shot classification results, and \cref{tab:domaingen-backbones} reports the results of the domain generalization task.
\tabZeroShotBackbones
\tabDomainGenBackbones

\subsection{Base-to-novel Generalization}
\label{app:base-to-novel}
\Cref{tab:base2novel-full} reports the full base-to-novel generalization results. TTC consistently improves both base-class and novel-class accuracy. Compared with CoOp, it increases the average base and novel accuracies by 1.60\,pp and 3.31\,pp, respectively.
The gain is larger on novel classes, which indicates that candidate verification mitigates the bias of prompt-tuned models toward base classes. TTC also achieves the best average base, novel, and harmonic-mean accuracies among all compared methods.
\tabBaseToNovelFull

\subsection{Success and Failure Study}
\label{app:success-and-failure}
\Cref{tab:taxonomy-per-task} reports success and failure cases for each setting. Across all settings, successes account for 78.14--84.75\% of the hit set, while overturning a correct initial prediction is the rarest case, occurring in only 2.39--4.46\%. Both failure cases have smaller median margins than successes. In particular, the median margin of overturning failures is 5.1--9.0$\times$ smaller than that of successes, indicating that TTC overturns correct predictions mainly when competing candidates are nearly tied.
\tabTaxonomyPerTask

\subsection{Divergence Metric Functions and Aggregated Verification Score}
\label{app:divergence-metric-functions-aggregated-verification-score}
\Cref{tab:metric-ablation-zs-full} reports the per-dataset zero-shot results for different divergence metrics. NSE achieves the best average accuracy with both backbones and outperforms cosine similarity on all 11 datasets with ResNet50 and on 10 of 11 datasets with ViT-B/16. In contrast, $l_2$ distance falls below the initial CLIP prediction on average, showing that the choice of divergence is critical for verification.
\tabMetricAblationZSFull

\Cref{tab:function-ablation-zs-full} reports the per-dataset zero-shot results for aggregated verification score variants. Every pair of components outperforms each of its components, and combining all three achieves the best average accuracy with both backbones, confirming that the terms provide complementary evidence.
\tabFunctionAblationZSFull

\section{Confidence Intervals for TTC Results}
\label{app:confidence-intervals}
We report 95\% confidence intervals for TTC results across three random seeds in \cref{tab:confidence-interval}. Since the ImageNet distribution-shift datasets are used only for domain generalization, we report their confidence intervals only in the `ZS/DG' column.
\tabConfidenceInterval

We also conduct a paired analysis where all methods share the same three stream permutations. \Cref{tab:paired-significance} reports the mean paired difference (Gap) with its 95\% confidence interval. Every Gap interval lies entirely above zero, and the Baseline CI is at most $\pm 0.74$, well below the gaps.
\tabPairedSignificance

\end{document}